\documentclass[sigconf]{acmart}

\usepackage{graphicx} 
\usepackage{subfigure}
\usepackage{balance}
\usepackage{multirow}
\usepackage{bbm}
\usepackage{color}
\usepackage{epstopdf}
\usepackage{hyperref}

\pdfoutput=1
\AtBeginDocument{%
  \providecommand\BibTeX{{%
    \normalfont B\kern-0.5em{\scshape i\kern-0.25em b}\kern-0.8em\TeX}}}

\begin{document}

\title{Counterfactually Measuring and Eliminating Social Bias in Vision-Language Pre-training Models}


\author{Yi Zhang}
\affiliation{%
  \institution{Beijing Jiaotong University, China}
  \state{}
  \country{}}
\email{yi.zhang@bjtu.edu.cn}

\author{Junyang Wang}
\affiliation{%
  \institution{Beijing Jiaotong University, China}
  \state{}
  \country{}}
\email{21120406@bjtu.edu.cn}

\author{Jitao Sang}
\affiliation{%
  \institution{Beijing Jiaotong University, China}
  \state{}
  \country{}}
\email{jtsang@bjtu.edu.cn}




\begin{abstract}
Vision-Language Pre-training (VLP) models have achieved state-of-the-art performance in numerous cross-modal tasks. Since they are optimized to capture the statistical properties of intra- and inter-modality, there remains risk to learn social biases presented in the data as well. In this work, we (1) introduce a counterfactual-based bias measurement \emph{CounterBias} to quantify the social bias in VLP models by comparing the [MASK]ed prediction probabilities of factual and counterfactual samples; (2) construct a novel VL-Bias dataset including 24K image-text pairs for measuring gender bias in VLP models, from which we observed that significant gender bias is prevalent in VLP models; and (3) propose a VLP debiasing method \emph{FairVLP} to minimize the difference in the [MASK]ed prediction probabilities between factual and counterfactual image-text pairs for VLP debiasing.
Although \emph{CounterBias} and \emph{FairVLP} focus on social bias, they are generalizable to serve as tools and provide new insights to probe and regularize more knowledge in VLP models.

\end{abstract}



\maketitle
\section{Introduction}\label{sec1}
\begin{figure}[ht]
  \centering
  \includegraphics[width=0.99\linewidth]{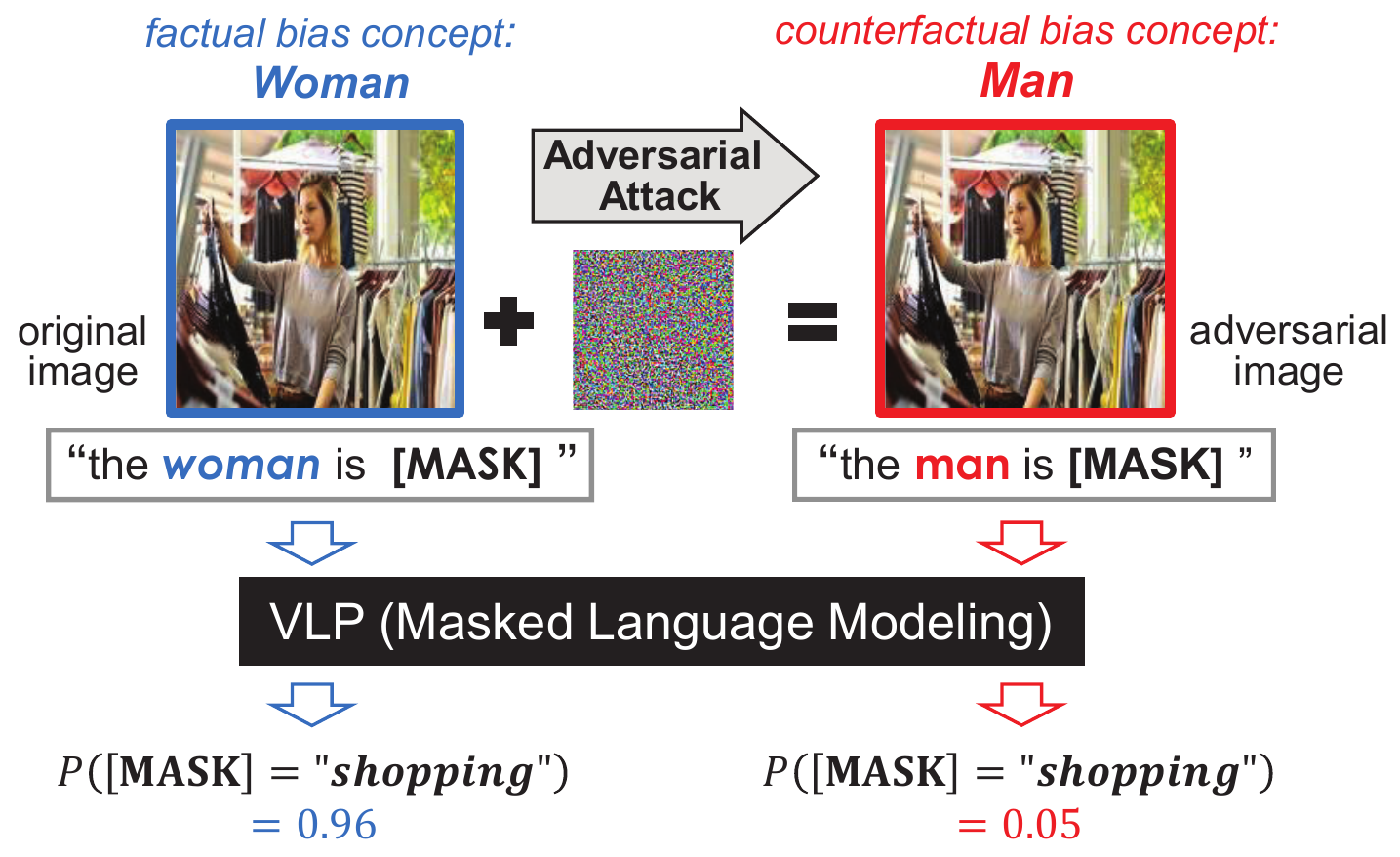}
   \setlength{\belowcaptionskip}{-0.1cm}
  \caption{We propose to measure social bias in VLP model by comparing the [MASK]ed prediction probabilities of factual and counterfactual samples.}
  \label{fig1}
\end{figure}


The past few years have witnessed the rapid development of Vision-Language Pre-training (VLP) models~\cite{chen2019uniter,li2020oscar,CuiROSITA,zhang2020devlbert}, and task-specific finetune on VLP models has become a new and state-of-the-art paradigm in many multimedia tasks~\cite{tan2019lxmert,lu2019vilbert,LuoCoCoBERT}.
Beyond accuracy, fairness which concerns about the discrimination towards socially protected or sensitive groups plays a critical role in trustworthy deployment of VLP models in downstream tasks. 
This is highly endorsed by many VLP model designers~\cite{li2021align}. Yet, very limited studies have begun to examine the fairness problem in VLP models, which typically apply the bias measurement designed for the unimodal pre-training (e.g., text-based language models~\cite{caliskan2017semantics}). 
The characteristics of VLP models regarding the interaction between visual modality and textual modality are largely ignored. 
This work is thus devoted to measuring and eliminating the social bias in VLP models by emphasizing/addressing its multimodal characteristics, with the goal of helping understand the existed social bias and inspiring the development of fair VLP models. 

Measuring the social bias in pre-training model essentially involves the correlation between bias and target concepts.
The cross-modal interaction of VLP models poses two main challenges for bias measurement. 
The first challenge is on modeling the bias and target concepts.
In VLP models, the joint modeling of visual modality delivers different embeddings w.r.t. image contexts, making direct calculating the distance in the embedding space as in~\cite{ross2021measuring} not applicable.
Inspired by the Masked Language Modeling (MLM) pre-training task commonly used in VLP models, we employ prompt-based query to probe the modeling of bias and target concepts in VLP models. As illustrated in Figure~\ref{fig1}, the concepts to be modeled are masked before issuing into VLP models. The resultant MLM prediction probability of [MASK]ed token $P(\text{[MASK]=``\emph{shopping}''})$ manifests VLP model's modeling of the concept given the input image-text pair. 

\begin{figure}[t]
  \centering
  \includegraphics[width=0.9\linewidth]{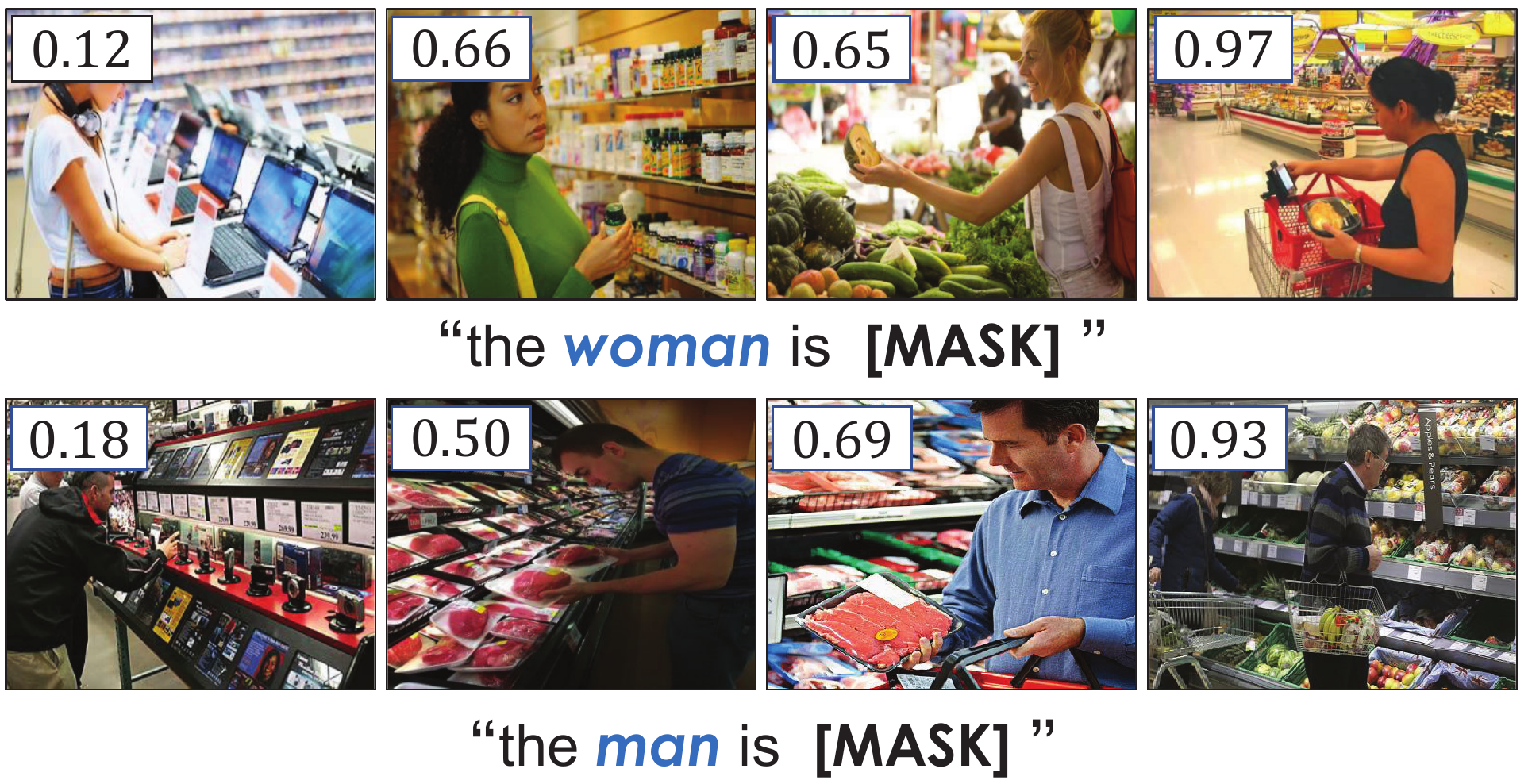}
  \setlength{\belowcaptionskip}{-5mm}
  \caption{Example ``shopping'' images with their predicted \emph{factual target probability} from ALBEF. The top and second row correspond to images with female and male bias concepts respectively. }
  \label{fig2}
\end{figure}

The second challenge is on quantifying the bias-target association. With MLM prediction probability for concept probing, a natural idea is to compare the prediction probabilities of the target concept in the observed image-text pairs of different bias concepts (e.g., the difference of $P(\text{[MASK]=``\emph{shopping}''})$ between \emph{male} and \emph{female} inputs.
However, relying on the observed data suffers from two problems: (1) There exists no adequate number of image pairs with only different bias concepts. For example,  when measuring the social bias of ``\emph{shopping}'' over gender, it is difficult to collect pairs of identical shopping images except for different genders. Other information beyond gender like the different background concepts can largely affect the MLM prediction probability, as detailed discussed in Section~\ref{sec3} for the example in Figure~\ref{fig2}. (2) The bias measurement result is highly sensitive to the observed dataset. Examining image-text datasets with different distributions will lead to inconsistent observations. 
To address this, we propose a counterfactual-based bias measure method (\textbf{\textit{CounterBias}}) to quantify the social bias in VLP models exempted from the limitation in observed data pairs. Specifically, we design a model-specific adversarial attack scheme to generate counterfactual images by altering the bias concept from factual images (e.g., attacking \emph{woman} to \emph{man} as illustrated in Figure~\ref{fig1}). This successfully alters the bias concept without changing other information. The resultant MLM prediction difference between factual and counterfactual image-text pairs thus manifests the social bias of the examined VLP models.

Since there exists no off-the-shelf dataset specially designed to analyze the social bias in VLP models, 
we propose VL-Bias, a dataset to facilitate the study on understanding model bias in VLP models. VL-Bias contains 52 activities and 13 occupations with a total of 24k image-text pairs. Using \emph{CounterBias}, we examined two typical VLP architectures, dual-stream and single-stream, on VL-Bias dataset. Several key observations include: (1) Social bias is prevalent in the examined three VLP models, and exists in both visual and language modalities; (2) The gender bias contained in VLP models is basically consistent with human gender stereotypes; (3) The examined single-stream ViLT and dual-stream ALBEF/TCL show different cross-modal conformities regarding the observed unimodal social bias; (4) The examined VLP models contain stronger gender bias than unimodal pre-training model BERT.

Inheriting the idea of measuring social bias, we propose a simple counterfactual-based method to eliminate the social bias in VLP models. Specifically, we first generate counterfactual samples of bias concepts in visual and language modalities, respectively. Then, the difference in the predicted [MASK]ed probability between factual and counterfactual image-text pairs is minimized, so that preventing the model from learning the association between bias and target concepts. Experimental results demonstrate the effectiveness in benchmark datasets.

Our contributions can be summarized as follows: 
\begin{itemize}
\item We propose to employ model-specific adversarial attacks on bias concepts to generate counterfactual samples, and counterfactually measure social bias at instance-level by comparing the [MASK]ed prediction probabilities of factual and counterfactual samples.
\item We introduce VL-Bias, a dataset for studying gender bias in VLP models. Observations are derived regarding typical VLP architectures, the consistency with human stereotypes, and the comparison with unimodal pre-training models. 
\item We propose a counterfactual bias elimination method, which eliminates bias by minimizing [MASK]ed probability discrepancy between factual and counterfactual samples.
\end{itemize}

\vspace{-0.5mm}
\section{Background and Related Works}

\subsection{Vision-Language Pre-training(VLP)}
Benefited from the cross-modal interactions performed by the transformer~\cite{vaswani2017attention}, Vision-Language Pre-training (VLP) has improved performance on various joint vision-and-language(V+L) downstream tasks~\cite{radford2021learning,hu2020vivo,lu2019vilbert,suhr2019corpus,jia2021scaling}. There are two mainstream architectures for bridging the cross-modal semantic gap: dual-stream architecture and single-stream architecture. The former first encodes visual and language modalities separately and then fuses the two representations through a transformer network; the latter collectively operates on a concatenation of image and text inputs. In the work, we consider ALBEF~\cite{li2021align} and TCL~\cite{yang2022vision} as example to examine the former, and ViLT~\cite{kim2021vilt} as example to examine the latter. 

\begin{figure*}[t]
  \centering
  \includegraphics[width=0.83\linewidth]{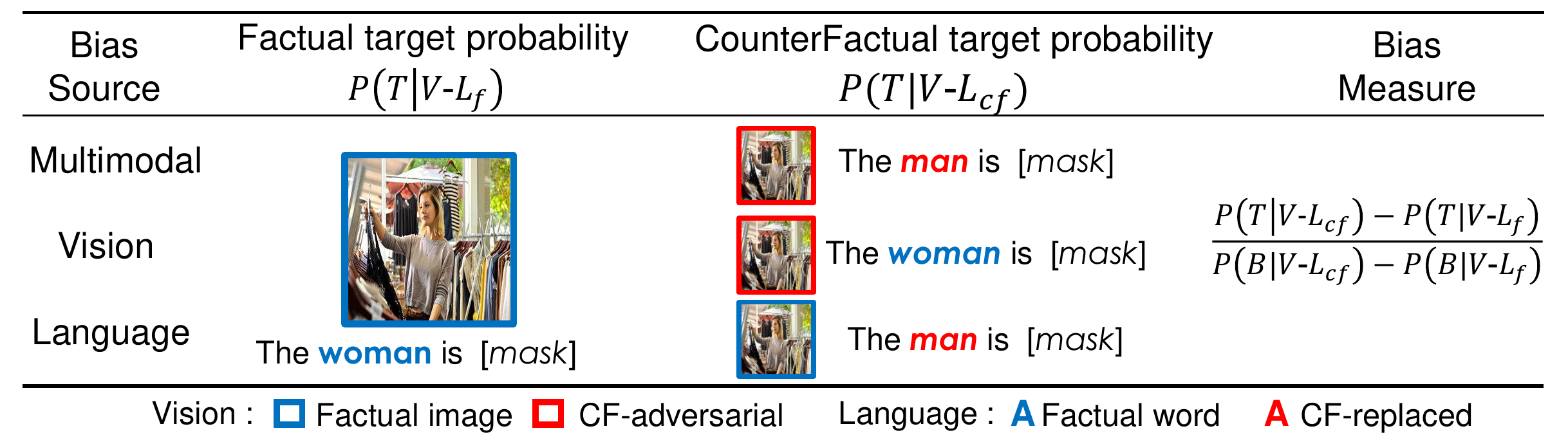}
  \caption{The proposed \emph{CounterBias} to counterfactually measure social bias between target concept $T$ and bias concept $B$.}
   \label{fig3}
\end{figure*}


\subsection{Bias Measurement and Elimination in VLP}
Much research work~\cite{zhao2019gender,caliskan2017semantics} has been done on evaluating bias in unimodal language models(LM). ~\cite{caliskan2017semantics} proposed cosine-based word embedding bias measure, and showed that word embeddings like GloVe~\cite{pennington2014glove} encode gender biases. Furthermore, ~\cite{srinivasan2021worst} investigates two unimodal biases in VL-BERT~\cite{su2019vl}.
However, there exists very limited work for such study in multimodal bias of VLP due to the cross-modal interaction. As far as we know, only~\cite{ross2021measuring} applied the cosine-based word embedding bias measure to multimodal bias of VLP models. However, ~\cite{kurita2019measuring} proved this method is not suitable for the contextual setting, as direct word will be affected by the context of image and text, resulting in inconsistent measure result. This motivates us to design bias measure scheme to consider the multimodal nature of the input in VLP models. For bias elimination, there has been works~\cite{sun2019mitigating,webster2020measuring} focused on eliminating bias in unimodal language models. However, eliminating the social bias in VLP models remain unexplored till now.


\section{Counterfactually Measuring Social Bias in VLP}\label{sec3}
The pre-training task of Masked Language Modeling (MLM) has been recognized as confidentially manifesting VLP model behavior by maximizing the mutual information(MI) between [MASK]ed token and the image-text input~\cite{li2021align}, and ~\cite{kurita2019measuring,srinivasan2021worst} utilized [MASK]ed token to probe the model's response of its context to measure unimodal contextual bias.
These provide theorectical basis and inspire us to use the prediction of [MASK]ed token to probe VLP model's understanding of the bias and target concepts. 
Specifically, when we construct the input to VLP model with image and text template masking off certain target concept $T$ (e.g., template of ``the \emph{woman} is [MASK]'' masking off the activity concept ``\emph{shopping}'' in Figure~\ref{fig1}), the output prediction probability can indicate how VLP model probes the concept from the given factual image-text pair. We denote the calculation of \emph{factual target probability} as follows:
\begin{equation}
P(T\mid V\text{-}L_{f})= P(\text{[MASK]} = T|V_{f},L_{f})
\end{equation}

Social bias essentially refers to how the change of bias concept in the input influences the model's inference on the target concept. Note that ideally only the involved bias concept is changed but fixing all the other information, so as to accurately examine the inherent influence from changed bias concept to target concept inference. However, as discussed in Section~\ref{sec1}, it is almost impossible to perfectly control and keep the other variables unchanged. Figure~\ref{fig2} shows some ``\emph{shopping}'' images with their predicted \emph{factual target probability}. We can see among images with the same bias concept (in the same row), the diversity of other variables like environment and background objects imposes obvious influence on the predicted target probability, even more significant than the bias concepts (between the two rows). This indicates relying on the observed data for bias measurement is unreliable. To avoid the negative effects of image diversity as a confounder, we propose \emph{CounterBias} to counterfactually change the bias concepts from observed image-text pairs without changing other information, and quantify the social bias by examining the difference between the predicted target probability of observed factual data $P(T\mid V\text{-}L_{f})$ and generated counterfactual data $P(T\mid V\text{-}L_{{cf}})$. 


Regarding how to counterfactually change the bias concepts in the visual modality, a direct idea is to use GANs to edit images. However, It is generally accepted that GANs-based editing can unintentionally change other information(e.g., image background)\cite{jain2022imperfect,jain2020study,grover2019bias,choi2020fair,menon2020pulse} , which defeats the purpose of controlling other information unchanged.
\vspace{2mm}

\noindent\textbf{Model-specific adversarial attacks.}\hspace{2mm} To avoid changing information other than bias concept, we propose model-specific adversarial attacks to counterfactually change the bias concepts in image.
In addition to affecting other information as little as possible, 
model-specific adversarial attacks ensure that all bias features that the VLP model relies on are counterfactually altered.
Specifically, we create a bias prompt template with masking the factual bias concept, “a [MASK] is in the picture”, as text input to guide the VLP model to classify bias concept $B$ (e.g., \emph{woman}) in the image input. Then we use the prediction probability $P(B \mid V_{f})$ of the [MASK]ed bias concept word to represent the bias concept that the model extracts from the factual image $V_{f}$:
\begin{equation}\label{bmeasure}
P(B\mid V_f)= P(\text{[MASK]=B}|V_{f}, \text{bias prompt}))
\end{equation}
where $B$ is the factual bias concept of input image.
Similar to traditional adversarial attacks~\cite{kurakin2016adversarial,goodfellow2014explaining,madry2017towards}, using cross-entropy loss between $P(B\mid V_f)$ and counterfactual goal $B'$, and search the adversarial perturbation $\delta$ based on gradient of VLP model so that bias concept of $V_{f}+\delta$ is close to $B'$:
\begin{equation}
\delta = \underset{\left\|\delta\right\| \leq \varepsilon }{\arg \min } \mathcal{L} \left(P(\text{[MASK]}|V_{f}+\delta, \text{bias prompt}), B'\right)
\end{equation}
where $B'$ represents counterfactual goals, if $B$ is \emph{woman}, $B'$ is \emph{man}.
And $V_{f}+\delta$ is generated counterfactual image $V_{cf}$.

Regarding how to counterfactually change bias concepts in language modality, we manually replace the bias concept word in factual text $L_{f}$, e.g. gendered word \emph{woman} is replaced by \emph{man}. The generated counterfactual text is denoted as $L_{cf}$. 


Note that the bias concept is continuous in images rather than discrete, e.g., the intensity of gender features in images of \emph{woman} is continuous, so the change of $P(B|V\text{-}L)$ also needs to be considered in the measure of social bias.
The bias concepts in factual and counterfactual images, $P(B\mid V_f)$ and $P(B\mid V_{cf})$, are formulated as the predicted probability of the [MASK]ed biased concept word, similar to Equation~\ref{bmeasure}.
Considering that the bias concept in text is binarized, $P(B\mid L_f)$ can thus be directly defined as $1$, and $P(B\mid L_{cf})$ is $0$. The bias concept $P(B\mid V,L)$ in the image-text pairs can be formulated as the average of the $P(B\mid V)$ and $P(B\mid L)$. 

Once \emph{factual probability} and \emph{counterfactual probability} in target concept and bias concept are calculated, we can measure how the counterfactual change of bias concept influences the model's inference on the target concept, as summarized in Figure~\ref{fig3}.
Specifically, the involved social bias between bias concept $B$ and target concept $T$ can be measured as:
\begin{equation}
bias(T,B) = \frac{P(T\mid V\text{-}L_{cf})-P(T\mid V\text{-}L_f)}{P(B\mid V\text{-}L_{cf})-P(B\mid V\text{-}L_f)}
\end{equation}

Considering
the multi-modal characteristic of VLP, we are interested in measuring social bias both in multimodality,
and also in visual and language unimodality.
The difference between measuring multimodal and unimodal biases lies in the counterfactual samples used, i.e., the difference in $P(T\mid V\text{-}L_{cf})$ and $P(B\mid V\text{-}L_{cf})$ calculations.
The following two subsections described measuring multimodal bias and unimodal bias in detail.

\subsection{Measuring Multimodal Bias}
Multimodal Bias refers to how the change of bias concept in both the visual and language modalities influences the model's inference on the target concept. 
Given counterfactual image-text pair $(V_{cf},L_{cf})$ as input,
we denote the calculation of \emph{counterfactual target probability} as follows:
\begin{equation}\label{cc}
P(T\mid V_{cf},L_{cf})= P(\text{[MASK]} = T|V_{cf},L_{cf})
\end{equation}
The multimodal bias $bias_{V,L}(T,B) $ can be formulated as:
\begin{equation}
\label{b_vl}
bias_{V\text{-}L}(T,B) = \frac{P(T\mid V_{cf},L_{cf})-P(T\mid V\text{-}L_f)}{P(B\mid V_{cf},L_{cf})-P(B\mid V_{f},L_{f})}
\end{equation}


\noindent$bias_{V,L}(T,B)$ is an instance-level bias measure for model bias. Positive values indicate target concept $T$ biased towards $B$, and negative values indicate that the bias direction of $T$ is opposite to $B$. Further, we also measure dataset-level social bias, by applying indicator function to align the bias directions represented by positive values, and calculate the average bias over a set of instances $S_T$ which contain the specific target concept $T$:
\begin{equation}
bias_{V\text{-}L}(T)=\mathbb{E}_{S_T}[\mathbbm{1}_{B=B_0}bias_{V\text{-}L}(T,B)-\mathbbm{1}_{B=B_1}bias_{V\text{-}L}(T,B)] 
\label{indicator}
\end{equation}
where $B_0$ and $B_1$ are pre-defined, and the sign of $bias_{V\text{-}L}(T)$ indicates the bias direction - positive for $B_0$, and negative for $B_1$. If $B_0$ is \emph{male}, positive values indicate $T$ biased towards \emph{male}.

\subsection{Measuring Unimodal Bias}
Unimodal bias refers to how the change of bias concept in unimodality influences the model's inference on the target concept.
We use counterfactual images and counterfactual texts to discover visual unimodal bias and language unimodal bias, respectively.

For visual unimodal bias, we calculate the \emph{counterfactual target probability} from given counterfactual visual $V_{cf}$ input and factual language input $L_{f}$ as follows:
\begin{equation}
P(T\mid V_f,L_{cf})= P(\text{[MASK]} = T|V_{cf},L_{f})
\end{equation}

\noindent Then, visual bias $bias_{V}(T,B)$ can be calculated as follows:
\begin{equation}
bias_{V}(T,B) = \frac{P(T\mid V_{cf},L_f)-P(T\mid V\text{-}L_f)}{P(B\mid V_{cf})-P(B\mid V_f)}
\end{equation}

For the language unimodal bias, we calculate the \emph{counterfactual target probability} from given counterfactual visual $V_{f}$ input and factual language input $L_{cf}$ as follows:
\begin{equation}
P(T\mid V_{f},L_{cf})= P(\text{[MASK]} = T|V_{f},L_{cf})
\end{equation}

\noindent Then, language bias $bias_{L}(T,B)$ can be calculated as follows:
\begin{equation}
bias_{L}(T,B) = \frac{P(T\mid V_{f},L_{cf})-P(T\mid V\text{-}L_f)}{P(B\mid L_{cf})-L(B\mid V_f)}
\end{equation}



\begin{figure}[t]
  \centering
  \includegraphics[width=0.95\linewidth]{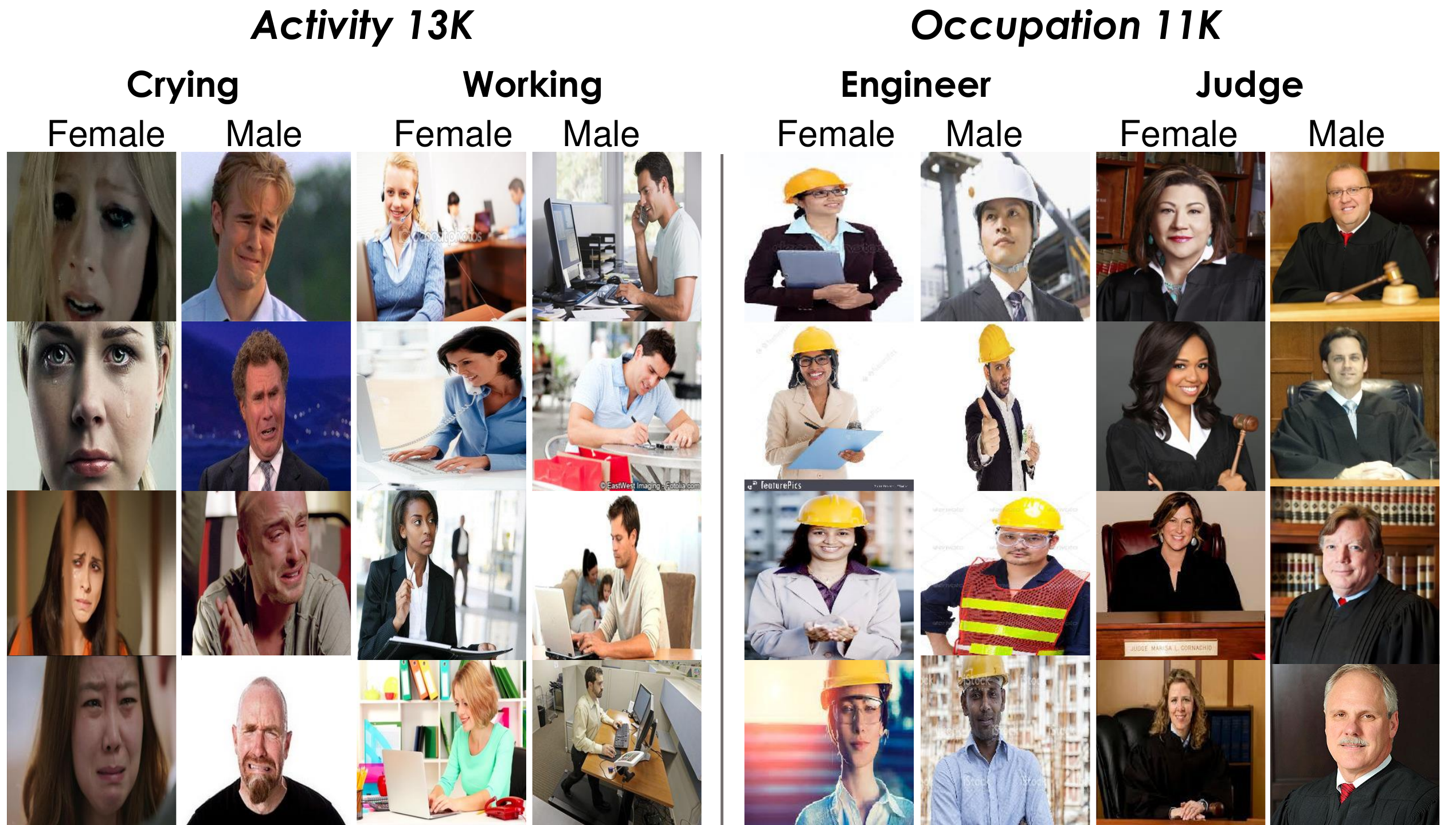}
  \caption{Example images in the collected VL-Bias dataset.}
  \label{dataset}
\end{figure}
\begin{table}[t] 
  \setlength{\abovecaptionskip}{1pt}

  \caption{Templates for caption generation.}
  \label{template}
  	\resizebox{0.47\textwidth}{!}
{
  \begin{tabular}{l|l}
    \toprule
    Type &Template \\
    \midrule
    Sentence w/o specification &The \{gender\} is [MASK]
\\
    Sentence w/ specification&The \{gender\} in the photo is [MASK]
\\
    Phrase w/o specification &A \{gender\} who is [MASK]
\\
    Phrase w/ specification& A photo of a \{gender\} who is [MASK]
\\
  \bottomrule

\end{tabular}
  }
\end{table}

\begin{figure*}[t!]
  \centering
  \includegraphics[width=0.98\linewidth]{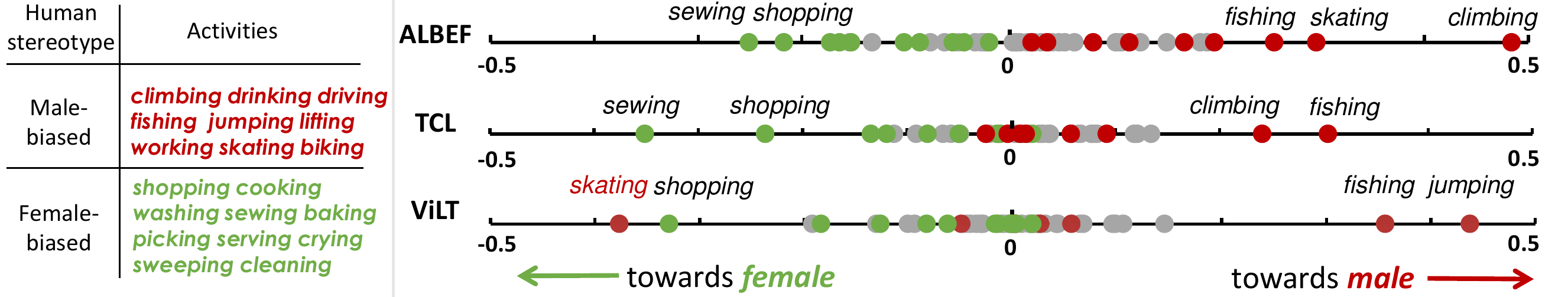}
  \caption{The human gender stereotypes (Left) and the observed multimodal bias (B@V-L) of three examined VLP models on different activities (Right).}
  \label{humuanste}
\end{figure*}

\vspace{-2mm}
\section{Measuring Results}~\label{biasresult}
\vspace{-6mm}
\subsection{VL-Bias Dataset}
Fairness research in CV mainly focuses on human face analysis~\cite{zhang2020towards,wang2019racial,huang2021trustworthy}, since existing face datasets such as CelebA~\cite{liu2015faceattributes} and UTKface~\cite{zhifei2017cvpr} have detailed annotation of both bias and target concepts of faces. For fairness in NLP models, researchers can measure the social bias of NLP models by means of manually constructing language templates~\cite{sun2020automatic}.
However, in vision-language research bridging CV and NLP, few studies focus on fairness problems due to the lack of both bias and target concepts annotation.

To address this issue, we construct VL-Bias to facilitate the study on analyzing social bias in VLP models. 
Using gender as the bias concept example, we selected 52 activities and 13 occupations related to humans\footnote{The VL-Bias dataset, as well as the complete activity and occupation concept list, is available at \href{url}{https://github.com/VL-Bias/VL-Bias} }.

For image collection, in addition to downloading human-related images from the Internet, we also collected human-related images from the existing image datasets, such as MS-COCO~\cite{lin2014microsoft}, Flickr~\cite{young2014image}. Then, we annotated gender concepts, activity concepts, and occupation concepts of the images and removed the images that did not explicitly contain these concepts.
We illustrate some annotated examples in Figure~\ref{dataset}.

We also generate corresponding captions for each image according to the annotation. 
Consider different languages and grammars have different effects on the prompt of [MASK], we use four templates~\footnote{We considered sentence and phrase, and whether to use "A photo of" in the text to specify that the text is about the image~\cite{radford2021learning}.}
described in Table~\ref{template} to generate captions.
Finally, for each template, we have collected 24k image-text pairs, including 13K for the 52 activities and 11K for the 13 occupations.

\begin{figure}[t]
  \centering
  \includegraphics[width=0.98\linewidth]{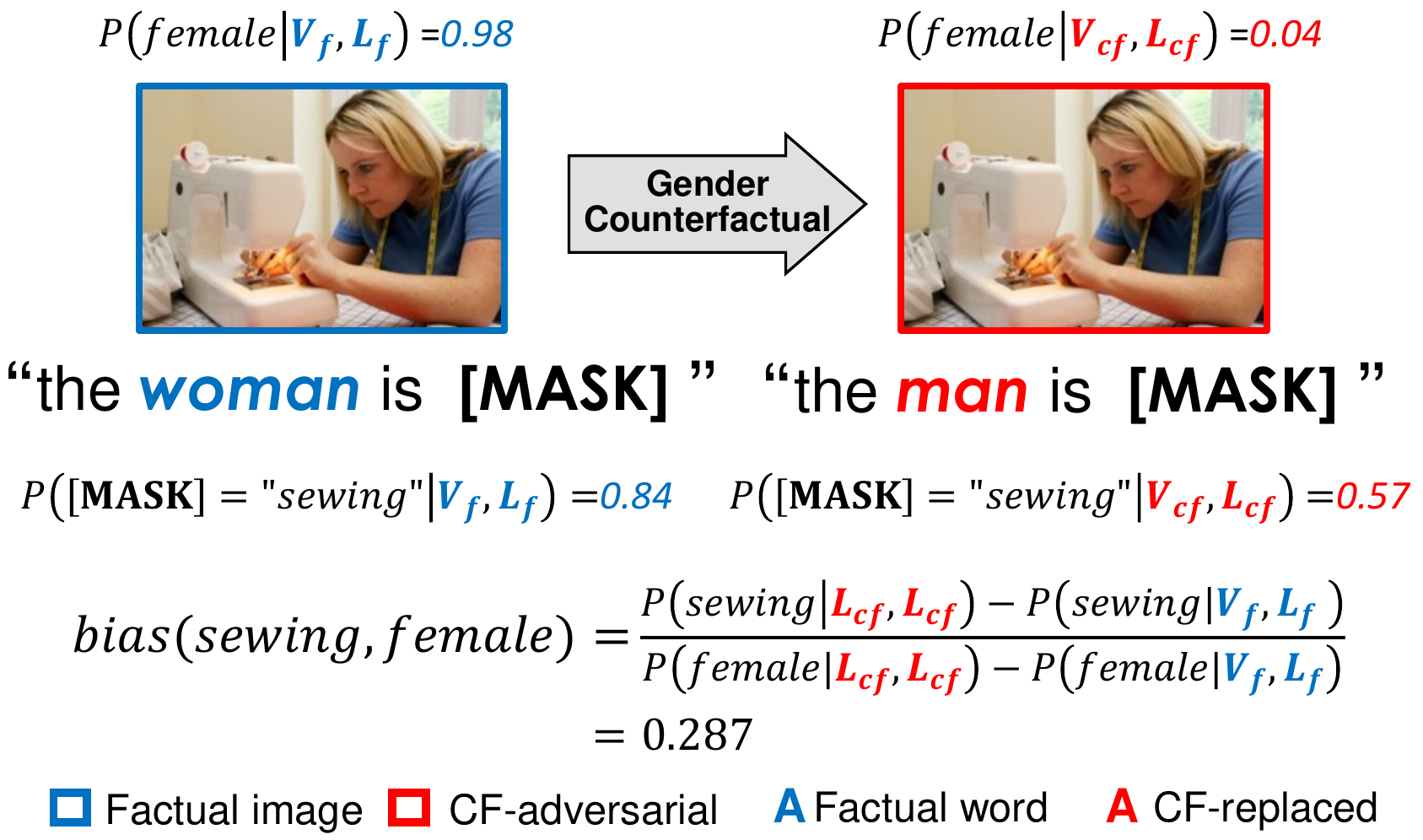}
  \caption{Example of the calculated social bias in VLP model ALBEF.}
  \vspace{-0.2cm}
  \label{exp1}
\end{figure}

\subsection{The Observed Social Bias in VLP}
Current VLP models can be divided into dual- and single- stream models according to architectures~\cite{kim2021vilt}. In dual-stream VLP models, the visual and textual inputs are first processed by two uni-modal encoders respectively, then the resulting representations are fed into cross-modal Transformer layers. For single-stream, VLP models use the concatenation of the image-text pair directly as the input of cross-modal Transformer layers. To deeply analyze the social bias in VLP models, we selected the representative VLP models of these two architectures respectively. For dual-stream, we consider ALBEF and TCL, and for single-stream, we consider ViLT. 
Taking gender bias as an example, we evaluate each VLP model on our VL-Bias dataset. 

Using the multimodal bias measurement as example, given target concept $T$ and bias concept $B$, instance-level social bias can be calculated according to Equation~\ref{b_vl}. One example is illustrated in Figure~\ref{exp1}. We can see the instance-level measurement provides intuitive understanding of the involved social bias. By aggregating instances with the same target concept, we can derive the final dataset-level social bias according to Equation~\ref{indicator}. The following elaborates some of the key observations.

\begin{figure*}[t]
  \centering
  \includegraphics[width=0.96\linewidth]{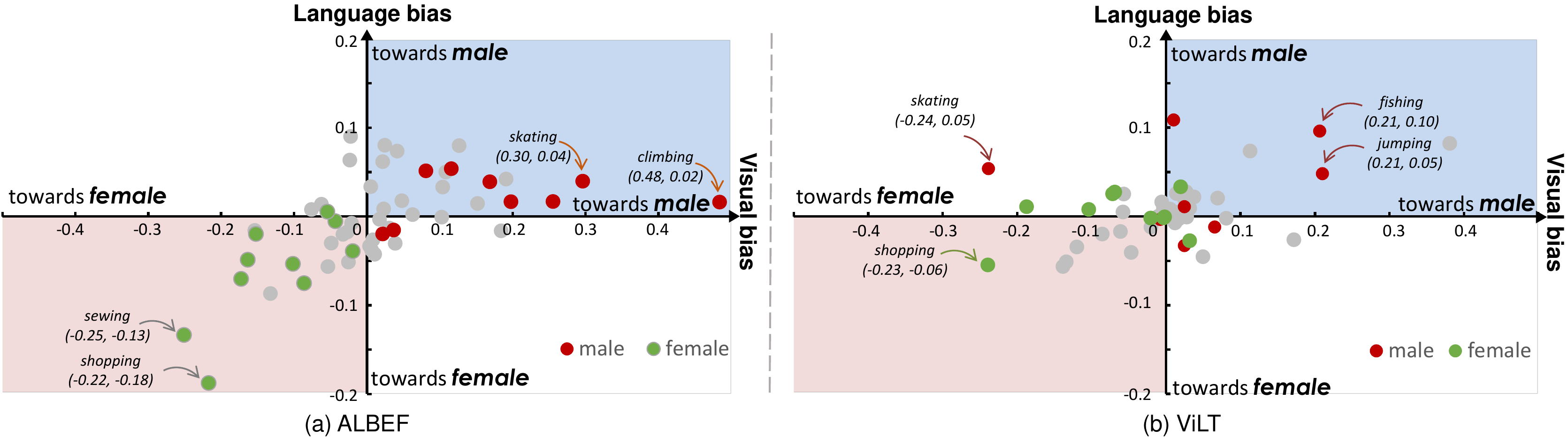}
  \caption{The visual modal bias (B@V) and language modal bias (B@L) in dual-stream ALBEF and single-stream ViLT.}
  \label{albef-vilt}
\end{figure*}


\begin{table}[t]
  \caption{The social bias (in $\%$) in VLP models in terms of multi-modal bias(B@V-L), visual modal bias(B@V), and language modal bias(B@L). 
  For each bias, we used four text templates to measure bias and calculate the average(Avg.).}
  
  
	\resizebox{0.46\textwidth}{!}
{
	\begin{tabular}	{l  l |  c  c  c | c  c  c  }
		\toprule
	 & & \multicolumn{3}{c|}{Activity 13K} &\multicolumn{3}{c}{Occupation 11K}\\

	 &&B@V-L &B@V&B@L& B@V-L &B@V&B@L \\
    \midrule

	\multirow{5}{*}{\rotatebox{90}{ALBEF}}  &S$_\text{W/O}$& 10.29 &9.44&3.98& 13.31 &9.62&2.38 \\
	 &S$_\text{W/}$& 11.33 &6.54&7.15& 16.00 &10.46&6.46 \\&P$_\text{W/O}$ & 10.75 &8.21&4.27& 11.85 &7.85&1.69 \\ &P$_\text{W/}$ & 9.85 &7.54&3.31& 11.31 &8.23&2.08 \\ &Avg.& \textbf{10.56} &\textbf{7.93}&\textbf{4.68}& \textbf{13.12} &\textbf{9.04}&\textbf{3.15} \\
	     \midrule
	\multirow{5}{*}{\rotatebox{90}{TCL}}  &S$_\text{W/O}$& 7.62&6.96&3.67& 12.00 &10.85&3.92 \\
	&S$_\text{W/}$ & 8.94 &6.15&5.50& 16.46 &12.01&6.00 \\ &P$_\text{W/O}$& 7.52 &5.90&2.65& 10.15 &6.85&4.77 \\ &P$_\text{W/}$ & 6.88 &5.38&2.37& 8.38 &5.46&2.85 \\&Avg. & \textbf{7.74} &\textbf{6.10}&\textbf{3.55}& \textbf{11.75} &\textbf{8.79}&\textbf{4.84} \\
	     \midrule

	\multirow{5}{*}{\rotatebox{90}{ViLT}} &S$_\text{W/O}$ & 8.62 &6.52&2.37&16.15& 17.23 &3.62 \\
	 &S$_\text{W/}$&9.48& 8.71 &2.35&14.92& 17.35 &4.62 \\& P$_\text{W/O}$&14.65& 14.38 &2.27&12.92& 17.61 &3.38 \\ &P$_\text{W/}$ & 11.19 &11.85&1.69& 13.23 &15.15&4.46 \\ &Avg.& \textbf{10.99} &\textbf{10.37}&\textbf{2.17}& \textbf{14.31} &\textbf{16.84}&\textbf{4.02} \\

		\bottomrule
	\end{tabular}
 	}
%
	\label{main_result}
    \vspace{-0.7cm}
\end{table}

\vspace{2mm}
\noindent\textbf{Social bias is prevalent in the examined VLP models.}\hspace{1mm}
We report the gender bias of the 52 activities and 13 occupations in Table~\ref{main_result}, where each element represents the average absolute value of the measured bias results.
Main observations include: (1) all three VLP models involve with significant gender bias. Taking multimodal bias (B@V-L) of ALBEF in activities as an example, $10.56\%$ means that by altering the gender information of the input, the model's predicted probability of [MASK]ed activity will change by an average of $10.56\%$.
(2) Gender bias exists in both modalities of VLP model and is more serious in visual modality (B@V) than in language modality(B@L). 
We conjecture that this is because images have richer information than text, which leads the model to capture more statistical properties within the visual modality. This further leads to stronger social bias in visual modality than in language modality.
(3) Bias measurement is sensitive to the choice of text input template, and using the average(Avg. in Table~\ref{main_result}) of multiple templates as the measurement result is more generalized.


\vspace{2mm}
\noindent\textbf{The gender bias contained in VLP models is basically consistent to human gender stereotypes.}\hspace{1mm}
To investigate whether the social bias involved in VLP models matches that from human stereotypes, we recruited 100 Amazon Mechanical Turk workers for labeling the stereotypes for the 52 activities and retained 19 activities with agreed gendered stereotyped label~\footnote{~\small{
For each activity, we asked 100 distinct AMT workers to label it as \emph{male-biased}, \emph{female-biased} or\emph{ neutral}. An activity obtains gender-biased stereotypes label only when at least $90\%$ works offer the same label. The rest 33 activities are associated with neutral label in the resultant human stereotype. }}(\emph{male} or \emph{female}-biased, illustrated in Figure~\ref{humuanste} (Left)).

In Figure~\ref{humuanste} (Right), we plot the multimodal bias (B@V-L) of examined VLP models with each activity, marked with human stereotypes using colored points(red for male-biased, green for female-biased, gray for neutral).
The sign of B@V-L indicates the direction of gender bias, positive for “male,” and negative for “female”.
We observed that the social bias B@V-L manifests is basically consistent with human gender stereotypes, which indicates that VLP models inherit human stereotypes from the training data.
In the three examined VLP models, compared to ALBEF and TCL, ViLT slightly deviates from human stereotypes in some of the activities, such as \emph{"skating"}. This may be related to the inconsistency in the bias direction between visual and language modalities in the following analysis.

\vspace{2mm}
\noindent\textbf{The examined dual-stream ALBEF/TCL and single-stream ViLT show different cross-modal conformities regarding the observed unimodal social bias.}\hspace{1mm}
Unlike unimodal pre-training models, there are complex vision-language cross-modal interactions in VLP models. The relationship between the social bias in two modalities may be of interest.
Taking dual-stream ALBEF and single-stream ViLT as examples, in Figure~\ref{albef-vilt}, we use scatter plots to compare the difference between visual modal bias ($x$-axis) and language modal bias ($y$-axis). Each point corresponds to an activity, with three colors representing different human stereotypes.
For ALBEF (Figure~\ref{albef-vilt}(a)), it is easy to find the strong conformity between visual modal bias and language modal bias: all activities are basically located in the first and third quadrants of the coordinate system, which means that the discovered gender biases in two modalities follow the same direction. Moreover, the bias of two modalities in $84\%$ of activities is consistent with human stereotypes.
However, the bias in ViLT is different from ALBEF, as shown in Figure~\ref{albef-vilt}(b), visual modal bias and language modal bias have large differences in direction, whit activities widely spreading across the four quadrants. Bias in unimodality is also inconsistent with human stereotypes in some activities, such as visual bias in \emph{skating} is also inconsistent with humans.


The conformity between visual modal bias and language modal bias in dual-stream ALBEF is possibly attributed to the explicit align constraints between the two unimodal representations (i.e., Image-Text Contrastive Learning pre-training task). 
By aligning the two modalities in dual-stream, ALBEF forces the two modalities to learn only shared information between visual and textual inputs, resulting in more consistent social biases contained in the two modalities, and biases in both two modalities are also consistent with human stereotypes. 
Moreover, the inconsistency between the learned information of the two modalities in ViLT again explains the reason for the slight deviation of multimodal bias(B@V-L) and human stereotypes.

\begin{table}[t]

  \caption{The social bias in language modality of three VLP models and language model BERT.}
  	\label{vlps-bert}
  \setlength{\abovecaptionskip}{-0.5cm}
{
	\begin{tabular}	{l   |  c  c  c  c   }
		\toprule	 	
	 \multirow{1}{*}{Target concept} &  ALBEF &TCL&ViLT& BERT \\

	\midrule
	52 Activities & 0.53 & 0.59 & 0.39 & \textbf{0.15}\\
	13 Occupations  & 0.69 & 0.71 & 0.48 &\textbf{0.20}\\
		\bottomrule
	\end{tabular}
%
}
    \vspace{-0.5cm}
    
\end{table}		
\vspace{2mm}
\noindent\textbf{The examined VLP models contain stronger gender bias than unimodal pre-training model BERT.} \hspace{1mm}
It is common practice for VLP models to use BERT~\cite{devlin2018bert} as backbone. We investigate the difference in language modal bias between BERT-based VLP models and original BERT.

For a fair comparison, we remove visual input when predicting [MASK]ed target concepts, which ensures that both VLP models and BERT rely only on language input:
\begin{equation}
\begin{aligned}
&P(T\mid L_f)= P(\text{[MASK]} = T|V_\text{\small{none}},L_f)\\
&P(T\mid L_{cf})= P(\text{[MASK]} = T|V_\text{\small{none}},L_{cf})
\end{aligned}
\end{equation}
Then, considering that the absence of visual input may result in a lower value of $P(T\mid L_{f/cf})$, we use log probability for stable numerical computations:
\begin{equation}
\label{equlog}
bias_L(T,B) = \frac{\log P(T\mid L_{cf})-\log P_L(T\mid L_f)}{P(B\mid L_{cf})-P(B\mid L_f)}
\end{equation}

For both VLP models and BERT, we use Equation~\ref{equlog} to calculate language modal bias. We report the language modal bias of the 52 activities and 13 occupations in Table~\ref{vlps-bert}. The results show that the examined three VLP models contain stronger gender bias than unimodal pre-training model BERT. This suggests that learning of cross-modal representations generally exacerbates the dependence of social biases in language modality.


\begin{figure}[t]
  \centering
  \includegraphics[width=0.9\linewidth]{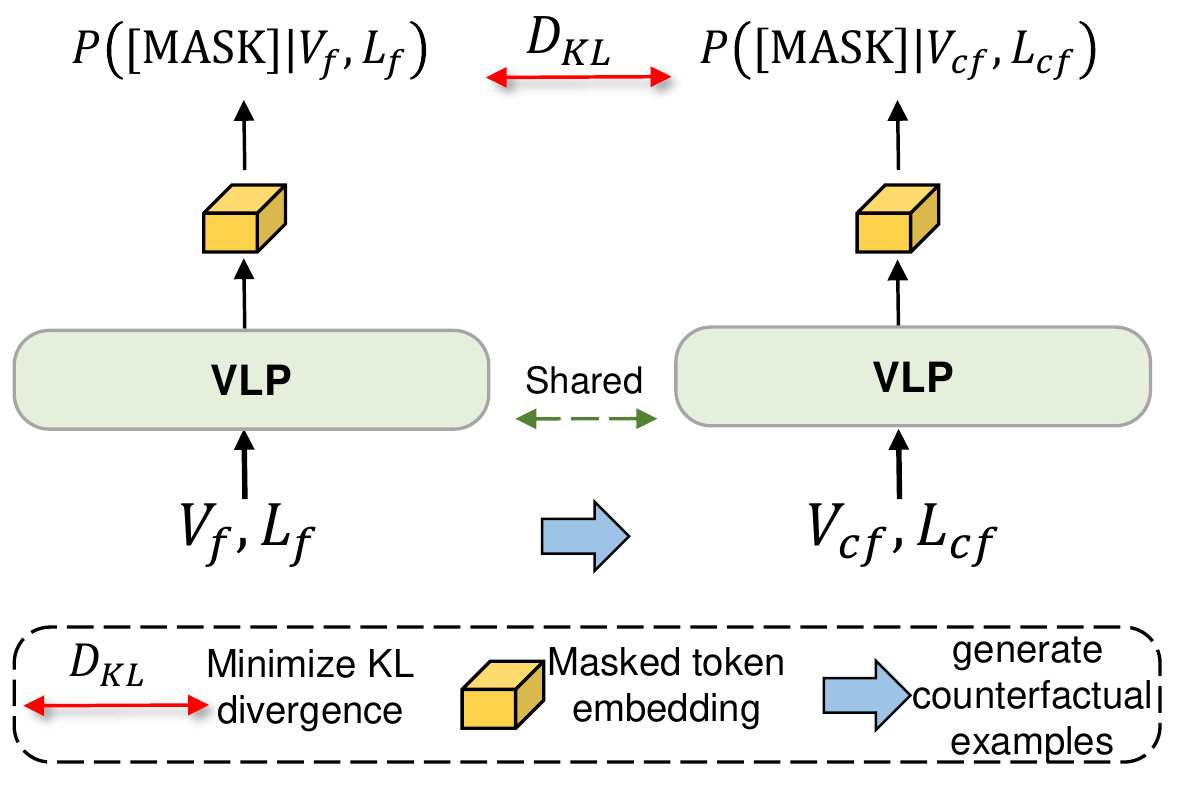}
   \setlength{\belowcaptionskip}{-0.5cm}
  \caption{The overall framework of our proposed \emph{FairVLP}. 
}
  \label{FairVLP}
\end{figure}

\section{Counterfactually Eliminating Social Bias in VLP}

\subsection{Method}
Section~\ref{debias_result} observes that social bias is prevalent in the examined VLP models, which can lead to unintended social consequences in downstream tasks.
Inheriting the idea of counterfactually measuring social bias,
we propose a fair Vision-Language Pre-training task \emph{FairVLP} that eliminates social bias by minimizing the difference in the predicted [MASK]ed probability between factual and counterfactual image-text pairs. 



%



Following the setup of masked language modeling (MLM) pre-training tasks commonly used in VLP models, our \emph{FairVLP} is applicable to both single- and dual-stream structures without being restricted to a specific VLP architecture. For the image-text pairs input at each training step $i$, we randomly mask out the text input tokens with probability $q$ and replace them with the special token [MASK]. The image and [MASK]ed text are denoted as $V^i_f$ and $L^i_f$.

\noindent\textbf{Counterfactual samples generation.}\hspace{2mm} Similar to Section~\ref{sec3}, for factual image $V^i_f$, we use model-specific adversarial attack to generate counterfactual image $V^i_{cf}$ by altering the bias concept of $V_f$. And we directly replace the words about bias concept (e.g., \emph{man} is replaced to \emph{woman}) in factual text $L^i_f$ to generate counterfactual text $L^i_{cf}$. 

\noindent\textbf{Minimize the difference in the prediction of [MASK]ed words between factual and counterfactual samples.}\hspace{2mm}
Following the formulation of MLM, we can utilize both the image $V$ and the text $L$ to predict [MASK]ed words $P(\text{[MASK]} \mid V,L)$. 
As shown in Figure~\ref{FairVLP}, we use factual and counterfactual samples to predict $P(\text{[MASK]} \mid V^i_f,L^i_f)$ and $P(\text{[MASK]} \mid V^i_{cf},L^i_{cf})$.

At current training step, our \emph{FairVLP} method eliminates social bias in VLP models by minimizing the Kullback-Leibler(KL) divergence in [MASK]ed word prediction distributions between the factual and counterfactual image-text pairs:
\begin{equation}
\mathcal{L}_{KL}=\mathcal{D}_{KL}(P(\text{[MASK]} \mid V^i_f,L^i_f) \| P(\text{[MASK]} \mid V^i_{cf},L^i_{cf}))
\end{equation}
With the basic cross-entropy loss $\mathcal{H}$ for the factual and counterfactual samples in MLM task:
\begin{equation}
\begin{aligned}
\mathcal{L}_{\mathrm{mlm}}&= \mathcal{H}(\boldsymbol{M}^i, P(\text{[MASK]} \mid V^i_f,L^i_f)) \\
&+\mathcal{H}(\boldsymbol{M}^i, P(\text{[MASK]} \mid V^i_{cf},L^i_{cf}))
\end{aligned}
\end{equation}
where $\boldsymbol{M}^i$ is the [MASK]ed word, and the final training objective is to minimize $L_{FairVLP}$:
\begin{equation}
\mathcal{L}_{\emph{FairVLP}}=(1-\alpha)\mathcal{L}_{\mathrm{mlm}}+\alpha \cdot \mathcal{L}_{KL}
\end{equation}
where $\alpha$ is the coefficient weight to control $\mathcal{L}_{KL}$.

By iteratively generating counterfactual samples and minimizing $\mathcal{L}_{\emph{FairVLP}}$,  The generated counterfactual samples $V^i_{cf}$ and $L^i_{cf}$ are guaranteed to have counterfactual properties for the VLP model at training step $i$. Minimizing the difference in prediction between factual and counterfactual samples, as a result, the VLP model's inference on target concept is independent of the bias concept at training step $i$. Throughout pre-training process, \emph{FairVLP} consistently prevents the VLP from learning social bias.




\begin{table}[t] 
  \caption{Social bias(in $\%$) in VLP models with different debiasing pre-training tasks. }
  \label{debias_result}
  	\resizebox{0.47\textwidth}{!}
{
	\begin{tabular}	{l   |  c  c  c | c  c  c  }
		\toprule	 	
	 \multirow{2}{*}{Method} & \multicolumn{3}{c|}{Activity 13K} &\multicolumn{3}{c}{Occupation 11K}\\
	 	 &B@V-L &B@V&B@L& B@V-L &B@V&B@L \\
    \midrule
    	vanilla  & 13.02 &11.10&4.89&14.84& 14.23 &3.15 \\
    	GS & 11.21 &9.25&4.03&12.47& 12.35 &2.98 \\
    	DR & 11.17 &9.54&4.39&13.52& 12.19 &3.08 \\
    	CycleGAN & 10.92 &9.23&4.32&12.07& 11.35 &3.01 \\

    	    \midrule
    	
    	FairVLP-V  & 8.41 &7.23&3.91&9.11& 7.98 &2.86 \\
    	FairVLP-L  & 8.53 &7.46&3.82&11.23& 9.61 &1.77 \\
    	    \midrule
    	FairVLP  & \textbf{6.97} &\textbf{6.76}&\textbf{3.29}&\textbf{7.74}& \textbf{7.15} &\textbf{1.72} \\

  \bottomrule
\end{tabular}
}
\end{table}

\begin{table*}[t]
  \setlength{\abovecaptionskip}{6pt}
	\caption
	{
		Fine-tuned image-text retrieval results on Flickr30K and COCO datasets.
	}
	\label{downmain}
	\centering	
	\resizebox{0.84\textwidth}{!}
{
	\begin{tabular}	{l   |  c  c  c  c  c  c | c  c  c  c  c  c }
		\toprule	 	
	 \multirow{2}{*}{Method} & \multicolumn{6}{c|}{Flickr30K (1K test set)} & \multicolumn{6}{c}{MSCOCO (5K test set)} \\
	  &  \multicolumn{3}{c}{TR}& \multicolumn{3}{c|}{IR} &  \multicolumn{3}{c}{TR}& \multicolumn{3}{c}{IR}\\
	 \midrule
	& R@1 &R@5&R@10& R@1 &R@5&R@10& R@1 &R@5&R@10& R@1 &R@5&R@10\\
	Vanilla  & 90.8 & 98.6 &99.3 &79.4 &94.9 &97.2 &70.5 &89.9 &94.6 &53.5 &79.3 &87.2 \\
	FairVLP  &90.8 & 98.8 & 98.3 & 78.9 & 94.8 & 97.3 & 70.3 &90.3 & 94.9 & 53.6 & 79.3 & 87.1 \\
	\bottomrule
	\end{tabular}
	}
	\vspace{-0.7ex}

\end{table*}

\subsection{Experimetnal Results}
\noindent\textbf{Experiment Setup.} 
We follow dual-stream ALBEF architecture and common MLM and ITM (image-text matching) pre-training tasks to pretrain debiasing VLP model on the MS-COCO dataset.
To evaluate the debiasing performance of \emph{FairVLP}, we consider two common and effective debiasing methods in language models as baselines:
Gender Swapping(\emph{GS})~\cite{zhao2018gender} and Dropout Regularization(\emph{DR})~\cite{webster2020measuring}.
\emph{GS} swaps gendered words in text input for mitigating the correlation between gender and other concepts.
\emph{DR} eliminates model over-fitting to gender concepts by increasing dropout rate. 
We migrated the idea of \emph{GS} to a visual modality and edited the concept of face gender in the images using \emph{CycleGAN}~\cite{zhu2017unpaired}.
We also consider two variants of our \emph{FairVLP} for ablation study, eliminating bias in visual modality by using only visual counterfactual samples(\emph{FairVLP-V}) and eliminating bias in language modality by only using language counterfactual samples(\emph{FairVLP-L}). For all methods, the learning rate is 1e-4 with 0.02 weight decay, and the batch size is 256.

\vspace{2mm}
\noindent\textbf{Debiasing performance.}\hspace{2mm} Table~\ref{debias_result} summarizes the debiasing performance of different methods measured on the VL-Bias dataset.
Main observations include:
(1) All the proposed three settings of \emph{FairVLP} obtain remarkable debiasing performance than that of baselines. Among three settings, by simultaneously performing counterfactual debiasing on both two modalities, \emph{FairVLP} demonstrates the best performance in eliminating social bias. 
(2) \emph{SW} and \emph{CycleGAN} show very limited debiasing results due to the lack of minimizing the difference in prediction of [MASK]ed word between the original sample and gender-swapping sample.
(3) \emph{DR} underperforms \emph{SW}, we owe this result to the random dropout does not change the statistical relationship in the training data. 

Moreover, We observe that \emph{FairVLP-V}, which is designed to eliminate the bias of visual modality, not only achieves a significant debiasing effect in the visual modality, but also demonstrates a remarkable debiasing effect in language modality. And \emph{FairVLP-L} also performs debiasing for visual modality.
We conjecture that this is because gender conflict in image-text pairs is an implicit form of data augmentation, which effectively makes the modeling of other concepts independent of gender concept.







\vspace{2mm}
\noindent\textbf{Downstream task performance.}\hspace{2mm} For the downstream tasks of the pre-training model, we follow the most commonly used Image-Text Retrieval among V+L tasks for model performance evaluation.
We evaluate pre-trained models on the Flickr30K and COCO benchmarks, and fine-tune the pre-trained model using the training samples from each dataset. 

Table~\ref{downmain} report results on fine-tuned image-text retrieval, respectively. Our \emph{FairVLP} pre-training task achieves comparable if not better performance with \emph{Vanilla}. 
This suggests that \emph{FairVLP} can leverage reasonably constructed counterfactual samples to effectively eliminate social bias without affecting the model's learning of reliable knowledge. This phenomenon further indicates that
the performance of the downstream task can be compatible with social bias elimination. 



\begin{figure}[]
  \centering
  \includegraphics[width=0.99\linewidth]{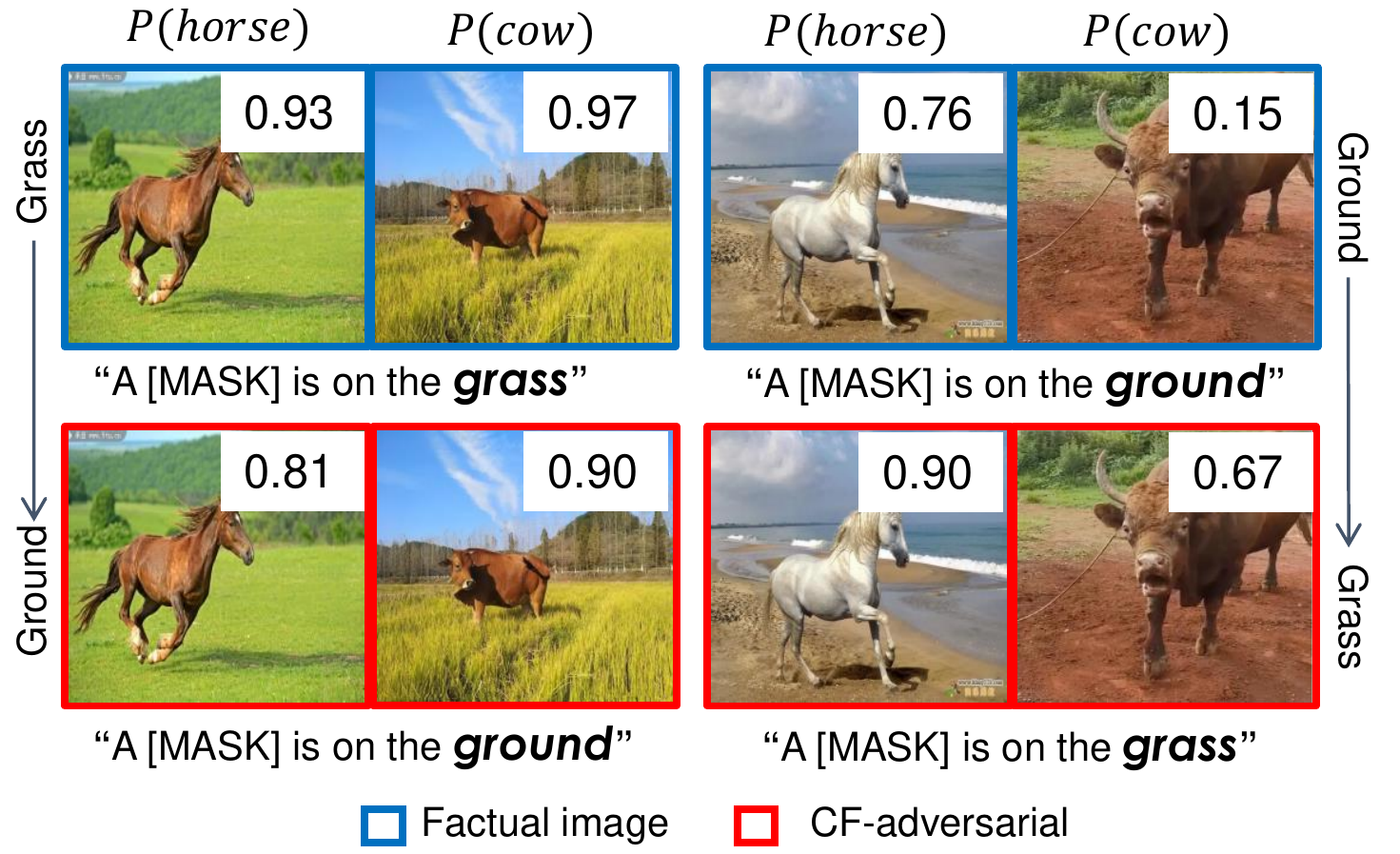}
  \setlength{\belowcaptionskip}{-0.5cm}
  \caption{Illustration of discovering spurious correlation by \emph{CounterBias}.}
  \label{discuss}
\end{figure}

\section{Discussion}
Visual-Language Pre-training models are designed to address the lack of downstream task-specific data, so the key to performance of downstream task is the learning of reliable cross-modal knowledge in pre-trained models.
In addition to revealing social bias in VLP models, we discuss that the proposed \emph{CounterBias} provides new insights into the opaque representation of knowledge in VLP models.
In this spirit, we examined the ability of \emph{CounterBias} to discover correlations between two arbitrary concepts in VLP models.
Specifically, taking ALBEF as an example, we use \emph{CounterBias} to measure the correlation between animal concepts (e.g., horse and cow) and background concepts (e.g., grass and ground).
As shown in Figure~\ref{discuss}, the MLM prediction difference between factual and counterfactual image-text pairs well reflects the correlation learned in ALBEF which shows that the model learns the spurious correlation between horse and grass, as well as cow and grass.
Accordingly, with available human annotation for spurious correlations, \emph{FairVLP} can also be considered as a method for regularizing the model to reliably learn knowledge.

\section{Conclusion}
In this work, we propose to measure social bias in VLP models by comparing how the model of target concept changes factual and counterfactual samples with different bias concepts. 
A novel VL-Bias dataset is introduced for studying social bias in multimodal models.
Taking gender as an example, our bias measure results on VLP models indicate that gender bias is prevalent in the examined VLP models. We propose to counterfactually eliminate social bias via minimizing the modeling differences between factual and counterfactual samples with different bias concepts. 
In the future, we are working towards counterfactually probing more knowledge as well as regulating the learning of VLP models.

\begin{acks}
This work is supported by the National Key R\&D Program of China (Grant No. 2018AAA0100604).
\end{acks}

\bibliographystyle{ACM-Reference-Format}
\balance
\bibliography{main}


\begin{thebibliography}{43}


\ifx \showCODEN    \undefined \def \showCODEN     #1{\unskip}     \fi
\ifx \showDOI      \undefined \def \showDOI       #1{#1}\fi
\ifx \showISBNx    \undefined \def \showISBNx     #1{\unskip}     \fi
\ifx \showISBNxiii \undefined \def \showISBNxiii  #1{\unskip}     \fi
\ifx \showISSN     \undefined \def \showISSN      #1{\unskip}     \fi
\ifx \showLCCN     \undefined \def \showLCCN      #1{\unskip}     \fi
\ifx \shownote     \undefined \def \shownote      #1{#1}          \fi
\ifx \showarticletitle \undefined \def \showarticletitle #1{#1}   \fi
\ifx \showURL      \undefined \def \showURL       {\relax}        \fi
\providecommand\bibfield[2]{#2}
\providecommand\bibinfo[2]{#2}
\providecommand\natexlab[1]{#1}
\providecommand\showeprint[2][]{arXiv:#2}

\bibitem[\protect\citeauthoryear{Caliskan, Bryson, and Narayanan}{Caliskan
  et~al\mbox{.}}{2017}]%
        {caliskan2017semantics}
\bibfield{author}{\bibinfo{person}{Aylin Caliskan}, \bibinfo{person}{Joanna~J
  Bryson}, {and} \bibinfo{person}{Arvind Narayanan}.}
  \bibinfo{year}{2017}\natexlab{}.
\newblock \showarticletitle{Semantics derived automatically from language
  corpora contain human-like biases}.
\newblock \bibinfo{journal}{\emph{Science}} \bibinfo{volume}{356},
  \bibinfo{number}{6334} (\bibinfo{year}{2017}), \bibinfo{pages}{183--186}.
\newblock


\bibitem[\protect\citeauthoryear{Chen, Li, Yu, Kholy, Ahmed, Gan, Cheng, and
  Liu}{Chen et~al\mbox{.}}{2020}]%
        {chen2019uniter}
\bibfield{author}{\bibinfo{person}{Yen-Chun Chen}, \bibinfo{person}{Linjie Li},
  \bibinfo{person}{Licheng Yu}, \bibinfo{person}{Ahmed~El Kholy},
  \bibinfo{person}{Faisal Ahmed}, \bibinfo{person}{Zhe Gan},
  \bibinfo{person}{Yu Cheng}, {and} \bibinfo{person}{Jingjing Liu}.}
  \bibinfo{year}{2020}\natexlab{}.
\newblock \showarticletitle{Uniter: Universal image-text representation
  learning}. In \bibinfo{booktitle}{\emph{ECCV}}.
\newblock


\bibitem[\protect\citeauthoryear{Choi, Grover, Singh, Shu, and Ermon}{Choi
  et~al\mbox{.}}{2020}]%
        {choi2020fair}
\bibfield{author}{\bibinfo{person}{Kristy Choi}, \bibinfo{person}{Aditya
  Grover}, \bibinfo{person}{Trisha Singh}, \bibinfo{person}{Rui Shu}, {and}
  \bibinfo{person}{Stefano Ermon}.} \bibinfo{year}{2020}\natexlab{}.
\newblock \showarticletitle{Fair generative modeling via weak supervision}. In
  \bibinfo{booktitle}{\emph{International Conference on Machine Learning}}.
  PMLR, \bibinfo{pages}{1887--1898}.
\newblock


\bibitem[\protect\citeauthoryear{Cui, Yu, Wang, Zhao, Zhang, Wang, and Yu}{Cui
  et~al\mbox{.}}{2021}]%
        {CuiROSITA}
\bibfield{author}{\bibinfo{person}{Yuhao Cui}, \bibinfo{person}{Zhou Yu},
  \bibinfo{person}{Chunqi Wang}, \bibinfo{person}{Zhongzhou Zhao},
  \bibinfo{person}{Ji Zhang}, \bibinfo{person}{Meng Wang}, {and}
  \bibinfo{person}{Jun Yu}.} \bibinfo{year}{2021}\natexlab{}.
\newblock \showarticletitle{ROSITA: Enhancing Vision-and-Language Semantic
  Alignments via Cross-and Intra-modal Knowledge Integration}. In
  \bibinfo{booktitle}{\emph{Proceedings of the 29th ACM International
  Conference on Multimedia}}. \bibinfo{pages}{797--806}.
\newblock


\bibitem[\protect\citeauthoryear{Devlin, Chang, Lee, and Toutanova}{Devlin
  et~al\mbox{.}}{[n.d.]}]%
        {devlin2018bert}
\bibfield{author}{\bibinfo{person}{Jacob Devlin}, \bibinfo{person}{Ming-Wei
  Chang}, \bibinfo{person}{Kenton Lee}, {and} \bibinfo{person}{Kristina
  Toutanova}.} \bibinfo{year}{[n.d.]}\natexlab{}.
\newblock \showarticletitle{Bert: Pre-training of deep bidirectional
  transformers for language understanding}.
\newblock  (\bibinfo{year}{[n.\,d.]}), \bibinfo{pages}{4171--4186}.
\newblock


\bibitem[\protect\citeauthoryear{Goodfellow, Shlens, and Szegedy}{Goodfellow
  et~al\mbox{.}}{2015}]%
        {goodfellow2014explaining}
\bibfield{author}{\bibinfo{person}{Ian~J. Goodfellow},
  \bibinfo{person}{Jonathon Shlens}, {and} \bibinfo{person}{Christian
  Szegedy}.} \bibinfo{year}{2015}\natexlab{}.
\newblock \showarticletitle{Explaining and harnessing adversarial examples}. In
  \bibinfo{booktitle}{\emph{3rd International Conference on Learning
  Representations}}.
\newblock


\bibitem[\protect\citeauthoryear{Grover, Song, Kapoor, Tran, Agarwal, Horvitz,
  and Ermon}{Grover et~al\mbox{.}}{2019}]%
        {grover2019bias}
\bibfield{author}{\bibinfo{person}{Aditya Grover}, \bibinfo{person}{Jiaming
  Song}, \bibinfo{person}{Ashish Kapoor}, \bibinfo{person}{Kenneth Tran},
  \bibinfo{person}{Alekh Agarwal}, \bibinfo{person}{Eric~J Horvitz}, {and}
  \bibinfo{person}{Stefano Ermon}.} \bibinfo{year}{2019}\natexlab{}.
\newblock \showarticletitle{Bias correction of learned generative models using
  likelihood-free importance weighting}.
\newblock \bibinfo{journal}{\emph{Advances in neural information processing
  systems}}  \bibinfo{volume}{32} (\bibinfo{year}{2019}).
\newblock


\bibitem[\protect\citeauthoryear{Hu, Yin, Lin, Wang, Zhang, Gao, and Liu}{Hu
  et~al\mbox{.}}{2020}]%
        {hu2020vivo}
\bibfield{author}{\bibinfo{person}{Xiaowei Hu}, \bibinfo{person}{Xi Yin},
  \bibinfo{person}{Kevin Lin}, \bibinfo{person}{Lijuan Wang},
  \bibinfo{person}{Lei Zhang}, \bibinfo{person}{Jianfeng Gao}, {and}
  \bibinfo{person}{Zicheng Liu}.} \bibinfo{year}{2020}\natexlab{}.
\newblock \showarticletitle{Vivo: Surpassing human performance in novel object
  captioning with visual vocabulary pre-training}.
\newblock  (\bibinfo{year}{2020}).
\newblock


\bibitem[\protect\citeauthoryear{Huang, Zhang, Zhang, Zhao, and Sang}{Huang
  et~al\mbox{.}}{2021}]%
        {huang2021trustworthy}
\bibfield{author}{\bibinfo{person}{Xiaowen Huang}, \bibinfo{person}{Jiaming
  Zhang}, \bibinfo{person}{Yi Zhang}, \bibinfo{person}{Xian Zhao}, {and}
  \bibinfo{person}{Jitao Sang}.} \bibinfo{year}{2021}\natexlab{}.
\newblock \showarticletitle{Trustworthy Multimedia Analysis}. In
  \bibinfo{booktitle}{\emph{Proceedings of the 29th ACM International
  Conference on Multimedia}}. \bibinfo{pages}{5667--5669}.
\newblock


\bibitem[\protect\citeauthoryear{Jain}{Jain}{2020}]%
        {jain2020study}
\bibfield{author}{\bibinfo{person}{Niharika Jain}.}
  \bibinfo{year}{2020}\natexlab{}.
\newblock \emph{\bibinfo{title}{A Study on Generative Adversarial Networks
  Exacerbating Social Data Bias}}.
\newblock \bibinfo{thesistype}{Ph.D. Dissertation}. \bibinfo{school}{Arizona
  State University}.
\newblock


\bibitem[\protect\citeauthoryear{Jain, Olmo, Sengupta, Manikonda, and
  Kambhampati}{Jain et~al\mbox{.}}{2022}]%
        {jain2022imperfect}
\bibfield{author}{\bibinfo{person}{Niharika Jain}, \bibinfo{person}{Alberto
  Olmo}, \bibinfo{person}{Sailik Sengupta}, \bibinfo{person}{Lydia Manikonda},
  {and} \bibinfo{person}{Subbarao Kambhampati}.}
  \bibinfo{year}{2022}\natexlab{}.
\newblock \showarticletitle{Imperfect ImaGANation: Implications of GANs
  exacerbating biases on facial data augmentation and snapchat face lenses}.
\newblock \bibinfo{journal}{\emph{Artificial Intelligence}}
  \bibinfo{volume}{304} (\bibinfo{year}{2022}), \bibinfo{pages}{103652}.
\newblock


\bibitem[\protect\citeauthoryear{Jia, Yang, Xia, Chen, Parekh, Pham, Le, Sung,
  Li, and Duerig}{Jia et~al\mbox{.}}{2021}]%
        {jia2021scaling}
\bibfield{author}{\bibinfo{person}{Chao Jia}, \bibinfo{person}{Yinfei Yang},
  \bibinfo{person}{Ye Xia}, \bibinfo{person}{Yi-Ting Chen},
  \bibinfo{person}{Zarana Parekh}, \bibinfo{person}{Hieu Pham},
  \bibinfo{person}{Quoc Le}, \bibinfo{person}{Yun-Hsuan Sung},
  \bibinfo{person}{Zhen Li}, {and} \bibinfo{person}{Tom Duerig}.}
  \bibinfo{year}{2021}\natexlab{}.
\newblock \showarticletitle{Scaling up visual and vision-language
  representation learning with noisy text supervision}. In
  \bibinfo{booktitle}{\emph{International Conference on Machine Learning}}.
  PMLR, \bibinfo{pages}{4904--4916}.
\newblock


\bibitem[\protect\citeauthoryear{Kim, Son, and Kim}{Kim et~al\mbox{.}}{2021}]%
        {kim2021vilt}
\bibfield{author}{\bibinfo{person}{Wonjae Kim}, \bibinfo{person}{Bokyung Son},
  {and} \bibinfo{person}{Ildoo Kim}.} \bibinfo{year}{2021}\natexlab{}.
\newblock \showarticletitle{Vilt: Vision-and-language transformer without
  convolution or region supervision}. In
  \bibinfo{booktitle}{\emph{International Conference on Machine Learning}}.
  PMLR, \bibinfo{pages}{5583--5594}.
\newblock


\bibitem[\protect\citeauthoryear{Kurakin, Goodfellow, and Bengio}{Kurakin
  et~al\mbox{.}}{2017}]%
        {kurakin2016adversarial}
\bibfield{author}{\bibinfo{person}{Alexey Kurakin}, \bibinfo{person}{Ian
  Goodfellow}, {and} \bibinfo{person}{Samy Bengio}.}
  \bibinfo{year}{2017}\natexlab{}.
\newblock \showarticletitle{Adversarial examples in the physical world}.
\newblock \bibinfo{journal}{\emph{5th International Conference on Learning
  Representations}} (\bibinfo{year}{2017}).
\newblock


\bibitem[\protect\citeauthoryear{Kurita, Vyas, Pareek, Black, and
  Tsvetkov}{Kurita et~al\mbox{.}}{2019}]%
        {kurita2019measuring}
\bibfield{author}{\bibinfo{person}{Keita Kurita}, \bibinfo{person}{Nidhi Vyas},
  \bibinfo{person}{Ayush Pareek}, \bibinfo{person}{Alan~W Black}, {and}
  \bibinfo{person}{Yulia Tsvetkov}.} \bibinfo{year}{2019}\natexlab{}.
\newblock \showarticletitle{Measuring Bias in Contextualized Word
  Representations}. In \bibinfo{booktitle}{\emph{Proceedings of the First
  Workshop on Gender Bias in Natural Language Processing}}.
  \bibinfo{pages}{166--172}.
\newblock


\bibitem[\protect\citeauthoryear{Li, Selvaraju, Gotmare, Joty, Xiong, and
  Hoi}{Li et~al\mbox{.}}{2021}]%
        {li2021align}
\bibfield{author}{\bibinfo{person}{Junnan Li}, \bibinfo{person}{Ramprasaath
  Selvaraju}, \bibinfo{person}{Akhilesh Gotmare}, \bibinfo{person}{Shafiq
  Joty}, \bibinfo{person}{Caiming Xiong}, {and} \bibinfo{person}{Steven
  Chu~Hong Hoi}.} \bibinfo{year}{2021}\natexlab{}.
\newblock \showarticletitle{Align before fuse: Vision and language
  representation learning with momentum distillation}.
\newblock \bibinfo{journal}{\emph{Advances in Neural Information Processing
  Systems}}  \bibinfo{volume}{34} (\bibinfo{year}{2021}).
\newblock


\bibitem[\protect\citeauthoryear{Li, Yin, Li, Zhang, Hu, Zhang, Wang, Hu, Dong,
  Wei, et~al\mbox{.}}{Li et~al\mbox{.}}{2020}]%
        {li2020oscar}
\bibfield{author}{\bibinfo{person}{Xiujun Li}, \bibinfo{person}{Xi Yin},
  \bibinfo{person}{Chunyuan Li}, \bibinfo{person}{Pengchuan Zhang},
  \bibinfo{person}{Xiaowei Hu}, \bibinfo{person}{Lei Zhang},
  \bibinfo{person}{Lijuan Wang}, \bibinfo{person}{Houdong Hu},
  \bibinfo{person}{Li Dong}, \bibinfo{person}{Furu Wei}, {et~al\mbox{.}}}
  \bibinfo{year}{2020}\natexlab{}.
\newblock \showarticletitle{Oscar: Object-semantics aligned pre-training for
  vision-language tasks}. In \bibinfo{booktitle}{\emph{European Conference on
  Computer Vision}}. Springer, \bibinfo{pages}{121--137}.
\newblock


\bibitem[\protect\citeauthoryear{Lin, Maire, Belongie, Hays, Perona, Ramanan,
  Doll{\'a}r, and Zitnick}{Lin et~al\mbox{.}}{2014}]%
        {lin2014microsoft}
\bibfield{author}{\bibinfo{person}{Tsung-Yi Lin}, \bibinfo{person}{Michael
  Maire}, \bibinfo{person}{Serge Belongie}, \bibinfo{person}{James Hays},
  \bibinfo{person}{Pietro Perona}, \bibinfo{person}{Deva Ramanan},
  \bibinfo{person}{Piotr Doll{\'a}r}, {and} \bibinfo{person}{C~Lawrence
  Zitnick}.} \bibinfo{year}{2014}\natexlab{}.
\newblock \showarticletitle{Microsoft coco: Common objects in context}. In
  \bibinfo{booktitle}{\emph{European conference on computer vision}}. Springer,
  \bibinfo{pages}{740--755}.
\newblock


\bibitem[\protect\citeauthoryear{Liu, Luo, Wang, and Tang}{Liu
  et~al\mbox{.}}{2015}]%
        {liu2015faceattributes}
\bibfield{author}{\bibinfo{person}{Ziwei Liu}, \bibinfo{person}{Ping Luo},
  \bibinfo{person}{Xiaogang Wang}, {and} \bibinfo{person}{Xiaoou Tang}.}
  \bibinfo{year}{2015}\natexlab{}.
\newblock \showarticletitle{Deep Learning Face Attributes in the Wild}. In
  \bibinfo{booktitle}{\emph{Proceedings of International Conference on Computer
  Vision (ICCV)}}.
\newblock


\bibitem[\protect\citeauthoryear{Lu, Batra, Parikh, and Lee}{Lu
  et~al\mbox{.}}{2019}]%
        {lu2019vilbert}
\bibfield{author}{\bibinfo{person}{Jiasen Lu}, \bibinfo{person}{Dhruv Batra},
  \bibinfo{person}{Devi Parikh}, {and} \bibinfo{person}{Stefan Lee}.}
  \bibinfo{year}{2019}\natexlab{}.
\newblock \showarticletitle{Vilbert: Pretraining task-agnostic visiolinguistic
  representations for vision-and-language tasks}.
\newblock \bibinfo{journal}{\emph{Advances in neural information processing
  systems}}  \bibinfo{volume}{32} (\bibinfo{year}{2019}).
\newblock


\bibitem[\protect\citeauthoryear{Luo, Li, Pan, Yao, Chao, and Mei}{Luo
  et~al\mbox{.}}{2021}]%
        {LuoCoCoBERT}
\bibfield{author}{\bibinfo{person}{Jianjie Luo}, \bibinfo{person}{Yehao Li},
  \bibinfo{person}{Yingwei Pan}, \bibinfo{person}{Ting Yao},
  \bibinfo{person}{Hongyang Chao}, {and} \bibinfo{person}{Tao Mei}.}
  \bibinfo{year}{2021}\natexlab{}.
\newblock \showarticletitle{CoCo-BERT: Improving Video-Language Pre-training
  with Contrastive Cross-modal Matching and Denoising}. In
  \bibinfo{booktitle}{\emph{Proceedings of the 29th ACM International
  Conference on Multimedia}}. \bibinfo{pages}{5600--5608}.
\newblock


\bibitem[\protect\citeauthoryear{Madry, Makelov, Schmidt, Tsipras, and
  Vladu}{Madry et~al\mbox{.}}{2017}]%
        {madry2017towards}
\bibfield{author}{\bibinfo{person}{Aleksander Madry},
  \bibinfo{person}{Aleksandar Makelov}, \bibinfo{person}{Ludwig Schmidt},
  \bibinfo{person}{Dimitris Tsipras}, {and} \bibinfo{person}{Adrian Vladu}.}
  \bibinfo{year}{2017}\natexlab{}.
\newblock \showarticletitle{Towards deep learning models resistant to
  adversarial attacks}.
\newblock \bibinfo{journal}{\emph{arXiv preprint arXiv:1706.06083}}
  (\bibinfo{year}{2017}).
\newblock


\bibitem[\protect\citeauthoryear{Menon, Damian, Hu, Ravi, and Rudin}{Menon
  et~al\mbox{.}}{2020}]%
        {menon2020pulse}
\bibfield{author}{\bibinfo{person}{Sachit Menon}, \bibinfo{person}{Alexandru
  Damian}, \bibinfo{person}{Shijia Hu}, \bibinfo{person}{Nikhil Ravi}, {and}
  \bibinfo{person}{Cynthia Rudin}.} \bibinfo{year}{2020}\natexlab{}.
\newblock \showarticletitle{Pulse: Self-supervised photo upsampling via latent
  space exploration of generative models}. In
  \bibinfo{booktitle}{\emph{Proceedings of the ieee/cvf conference on computer
  vision and pattern recognition}}. \bibinfo{pages}{2437--2445}.
\newblock


\bibitem[\protect\citeauthoryear{Pennington, Socher, and Manning}{Pennington
  et~al\mbox{.}}{2014}]%
        {pennington2014glove}
\bibfield{author}{\bibinfo{person}{Jeffrey Pennington},
  \bibinfo{person}{Richard Socher}, {and} \bibinfo{person}{Christopher~D
  Manning}.} \bibinfo{year}{2014}\natexlab{}.
\newblock \showarticletitle{Glove: Global vectors for word representation}. In
  \bibinfo{booktitle}{\emph{Proceedings of the 2014 conference on empirical
  methods in natural language processing (EMNLP)}}.
  \bibinfo{pages}{1532--1543}.
\newblock


\bibitem[\protect\citeauthoryear{Radford, Kim, Hallacy, Ramesh, Goh, Agarwal,
  Sastry, Askell, Mishkin, Clark, et~al\mbox{.}}{Radford et~al\mbox{.}}{2021}]%
        {radford2021learning}
\bibfield{author}{\bibinfo{person}{Alec Radford}, \bibinfo{person}{Jong~Wook
  Kim}, \bibinfo{person}{Chris Hallacy}, \bibinfo{person}{Aditya Ramesh},
  \bibinfo{person}{Gabriel Goh}, \bibinfo{person}{Sandhini Agarwal},
  \bibinfo{person}{Girish Sastry}, \bibinfo{person}{Amanda Askell},
  \bibinfo{person}{Pamela Mishkin}, \bibinfo{person}{Jack Clark},
  {et~al\mbox{.}}} \bibinfo{year}{2021}\natexlab{}.
\newblock \showarticletitle{Learning transferable visual models from natural
  language supervision}. In \bibinfo{booktitle}{\emph{International Conference
  on Machine Learning}}. PMLR, \bibinfo{pages}{8748--8763}.
\newblock


\bibitem[\protect\citeauthoryear{Ross, Katz, and Barbu}{Ross
  et~al\mbox{.}}{2020}]%
        {ross2021measuring}
\bibfield{author}{\bibinfo{person}{Candace Ross}, \bibinfo{person}{Boris Katz},
  {and} \bibinfo{person}{Andrei Barbu}.} \bibinfo{year}{2020}\natexlab{}.
\newblock \showarticletitle{Measuring Social Biases in Grounded Vision and
  Language Embeddings}.
\newblock \bibinfo{journal}{\emph{arXiv preprint arXiv:2002.08911}}.
\newblock


\bibitem[\protect\citeauthoryear{Srinivasan and Bisk}{Srinivasan and
  Bisk}{2021}]%
        {srinivasan2021worst}
\bibfield{author}{\bibinfo{person}{Tejas Srinivasan} {and}
  \bibinfo{person}{Yonatan Bisk}.} \bibinfo{year}{2021}\natexlab{}.
\newblock \showarticletitle{Worst of both worlds: Biases compound in
  pre-trained vision-and-language models}.
\newblock \bibinfo{journal}{\emph{arXiv preprint arXiv:2104.08666}}
  (\bibinfo{year}{2021}).
\newblock


\bibitem[\protect\citeauthoryear{Su, Zhu, Cao, Li, Lu, Wei, and Dai}{Su
  et~al\mbox{.}}{2019}]%
        {su2019vl}
\bibfield{author}{\bibinfo{person}{Weijie Su}, \bibinfo{person}{Xizhou Zhu},
  \bibinfo{person}{Yue Cao}, \bibinfo{person}{Bin Li}, \bibinfo{person}{Lewei
  Lu}, \bibinfo{person}{Furu Wei}, {and} \bibinfo{person}{Jifeng Dai}.}
  \bibinfo{year}{2019}\natexlab{}.
\newblock \showarticletitle{Vl-bert: Pre-training of generic visual-linguistic
  representations}.
\newblock \bibinfo{journal}{\emph{arXiv preprint arXiv:1908.08530}}
  (\bibinfo{year}{2019}).
\newblock


\bibitem[\protect\citeauthoryear{Suhr, Zhou, Zhang, Zhang, Bai, and Artzi}{Suhr
  et~al\mbox{.}}{2019}]%
        {suhr2019corpus}
\bibfield{author}{\bibinfo{person}{Alane Suhr}, \bibinfo{person}{Stephanie
  Zhou}, \bibinfo{person}{Ally Zhang}, \bibinfo{person}{Iris Zhang},
  \bibinfo{person}{Huajun Bai}, {and} \bibinfo{person}{Yoav Artzi}.}
  \bibinfo{year}{2019}\natexlab{}.
\newblock \showarticletitle{A Corpus for Reasoning about Natural Language
  Grounded in Photographs}. In \bibinfo{booktitle}{\emph{Proceedings of the
  57th Annual Meeting of the Association for Computational Linguistics}}.
  \bibinfo{pages}{6418--6428}.
\newblock


\bibitem[\protect\citeauthoryear{Sun, Gaut, Tang, Huang, ElSherief, Zhao,
  Mirza, Belding, Chang, and Wang}{Sun et~al\mbox{.}}{2019}]%
        {sun2019mitigating}
\bibfield{author}{\bibinfo{person}{Tony Sun}, \bibinfo{person}{Andrew Gaut},
  \bibinfo{person}{Shirlyn Tang}, \bibinfo{person}{Yuxin Huang},
  \bibinfo{person}{Mai ElSherief}, \bibinfo{person}{Jieyu Zhao},
  \bibinfo{person}{Diba Mirza}, \bibinfo{person}{Elizabeth Belding},
  \bibinfo{person}{Kai-Wei Chang}, {and} \bibinfo{person}{William~Yang Wang}.}
  \bibinfo{year}{2019}\natexlab{}.
\newblock \showarticletitle{Mitigating Gender Bias in Natural Language
  Processing: Literature Review}.
\newblock \bibinfo{journal}{\emph{Association for Computational Linguistics
  (ACL 2019)}} (\bibinfo{year}{2019}).
\newblock


\bibitem[\protect\citeauthoryear{Sun, Zhang, Harman, Papadakis, and Zhang}{Sun
  et~al\mbox{.}}{2017}]%
        {zhifei2017cvpr}
\bibfield{author}{\bibinfo{person}{Zeyu Sun}, \bibinfo{person}{Jie~M Zhang},
  \bibinfo{person}{Mark Harman}, \bibinfo{person}{Mike Papadakis}, {and}
  \bibinfo{person}{Lu Zhang}.} \bibinfo{year}{2017}\natexlab{}.
\newblock \showarticletitle{Age Progression/Regression by Conditional
  Adversarial Autoencoder}. In \bibinfo{booktitle}{\emph{IEEE Conference on
  Computer Vision and Pattern Recognition (CVPR)}}. IEEE.
\newblock


\bibitem[\protect\citeauthoryear{Sun, Zhang, Harman, Papadakis, and Zhang}{Sun
  et~al\mbox{.}}{2020}]%
        {sun2020automatic}
\bibfield{author}{\bibinfo{person}{Zeyu Sun}, \bibinfo{person}{Jie~M Zhang},
  \bibinfo{person}{Mark Harman}, \bibinfo{person}{Mike Papadakis}, {and}
  \bibinfo{person}{Lu Zhang}.} \bibinfo{year}{2020}\natexlab{}.
\newblock \showarticletitle{Automatic testing and improvement of machine
  translation}. In \bibinfo{booktitle}{\emph{Proceedings of the ACM/IEEE 42nd
  International Conference on Software Engineering}}.
  \bibinfo{pages}{974--985}.
\newblock


\bibitem[\protect\citeauthoryear{Tan and Bansal}{Tan and Bansal}{2019}]%
        {tan2019lxmert}
\bibfield{author}{\bibinfo{person}{Hao Tan} {and} \bibinfo{person}{Mohit
  Bansal}.} \bibinfo{year}{2019}\natexlab{}.
\newblock \showarticletitle{LXMERT: Learning Cross-Modality Encoder
  Representations from Transformers}. In \bibinfo{booktitle}{\emph{Proceedings
  of the 2019 Conference on Empirical Methods in Natural Language Processing
  and the 9th International Joint Conference on Natural Language Processing
  (EMNLP-IJCNLP)}}. \bibinfo{pages}{5100--5111}.
\newblock


\bibitem[\protect\citeauthoryear{Vaswani, Shazeer, Parmar, Uszkoreit, Jones,
  Gomez, Kaiser, and Polosukhin}{Vaswani et~al\mbox{.}}{2017}]%
        {vaswani2017attention}
\bibfield{author}{\bibinfo{person}{Ashish Vaswani}, \bibinfo{person}{Noam
  Shazeer}, \bibinfo{person}{Niki Parmar}, \bibinfo{person}{Jakob Uszkoreit},
  \bibinfo{person}{Llion Jones}, \bibinfo{person}{Aidan~N Gomez},
  \bibinfo{person}{{\L}ukasz Kaiser}, {and} \bibinfo{person}{Illia
  Polosukhin}.} \bibinfo{year}{2017}\natexlab{}.
\newblock \showarticletitle{Attention is all you need}.
\newblock \bibinfo{journal}{\emph{Advances in neural information processing
  systems}}  \bibinfo{volume}{30} (\bibinfo{year}{2017}).
\newblock


\bibitem[\protect\citeauthoryear{Wang, Deng, Hu, Tao, and Huang}{Wang
  et~al\mbox{.}}{2019}]%
        {wang2019racial}
\bibfield{author}{\bibinfo{person}{Mei Wang}, \bibinfo{person}{Weihong Deng},
  \bibinfo{person}{Jiani Hu}, \bibinfo{person}{Xunqiang Tao}, {and}
  \bibinfo{person}{Yaohai Huang}.} \bibinfo{year}{2019}\natexlab{}.
\newblock \showarticletitle{Racial faces in the wild: Reducing racial bias by
  information maximization adaptation network}. In
  \bibinfo{booktitle}{\emph{Proceedings of the ieee/cvf international
  conference on computer vision}}. \bibinfo{pages}{692--702}.
\newblock


\bibitem[\protect\citeauthoryear{Webster, Wang, Tenney, Beutel, Pitler,
  Pavlick, Chen, Chi, and Petrov}{Webster et~al\mbox{.}}{2020}]%
        {webster2020measuring}
\bibfield{author}{\bibinfo{person}{Kellie Webster}, \bibinfo{person}{Xuezhi
  Wang}, \bibinfo{person}{Ian Tenney}, \bibinfo{person}{Alex Beutel},
  \bibinfo{person}{Emily Pitler}, \bibinfo{person}{Ellie Pavlick},
  \bibinfo{person}{Jilin Chen}, \bibinfo{person}{Ed Chi}, {and}
  \bibinfo{person}{Slav Petrov}.} \bibinfo{year}{2020}\natexlab{}.
\newblock \showarticletitle{Measuring and reducing gendered correlations in
  pre-trained models}.
\newblock \bibinfo{journal}{\emph{arXiv preprint arXiv:2010.06032}}
  (\bibinfo{year}{2020}).
\newblock


\bibitem[\protect\citeauthoryear{Yang, Duan, Tran, Xu, Chanda, Chen, Zeng,
  Chilimbi, and Huang}{Yang et~al\mbox{.}}{2022}]%
        {yang2022vision}
\bibfield{author}{\bibinfo{person}{Jinyu Yang}, \bibinfo{person}{Jiali Duan},
  \bibinfo{person}{Son Tran}, \bibinfo{person}{Yi Xu}, \bibinfo{person}{Sampath
  Chanda}, \bibinfo{person}{Liqun Chen}, \bibinfo{person}{Belinda Zeng},
  \bibinfo{person}{Trishul Chilimbi}, {and} \bibinfo{person}{Junzhou Huang}.}
  \bibinfo{year}{2022}\natexlab{}.
\newblock \showarticletitle{Vision-Language Pre-Training with Triple
  Contrastive Learning}.
\newblock \bibinfo{journal}{\emph{Proceedings of the IEEE conference on
  computer vision and pattern recognition}} (\bibinfo{year}{2022}).
\newblock


\bibitem[\protect\citeauthoryear{Young, Lai, Hodosh, and Hockenmaier}{Young
  et~al\mbox{.}}{2014}]%
        {young2014image}
\bibfield{author}{\bibinfo{person}{Peter Young}, \bibinfo{person}{Alice Lai},
  \bibinfo{person}{Micah Hodosh}, {and} \bibinfo{person}{Julia Hockenmaier}.}
  \bibinfo{year}{2014}\natexlab{}.
\newblock \showarticletitle{From image descriptions to visual denotations: New
  similarity metrics for semantic inference over event descriptions}.
\newblock \bibinfo{journal}{\emph{Transactions of the Association for
  Computational Linguistics}}  \bibinfo{volume}{2} (\bibinfo{year}{2014}),
  \bibinfo{pages}{67--78}.
\newblock


\bibitem[\protect\citeauthoryear{Zhang, Jiang, Wang, Kuang, Zhao, Zhu, Yu,
  Yang, and Wu}{Zhang et~al\mbox{.}}{2020}]%
        {zhang2020devlbert}
\bibfield{author}{\bibinfo{person}{Shengyu Zhang}, \bibinfo{person}{Tan Jiang},
  \bibinfo{person}{Tan Wang}, \bibinfo{person}{Kun Kuang},
  \bibinfo{person}{Zhou Zhao}, \bibinfo{person}{Jianke Zhu},
  \bibinfo{person}{Jin Yu}, \bibinfo{person}{Hongxia Yang}, {and}
  \bibinfo{person}{Fei Wu}.} \bibinfo{year}{2020}\natexlab{}.
\newblock \showarticletitle{Devlbert: Learning deconfounded visio-linguistic
  representations}. In \bibinfo{booktitle}{\emph{Proceedings of the 28th ACM
  International Conference on Multimedia}}. \bibinfo{pages}{4373--4382}.
\newblock


\bibitem[\protect\citeauthoryear{Zhang and Sang}{Zhang and Sang}{2020}]%
        {zhang2020towards}
\bibfield{author}{\bibinfo{person}{Yi Zhang} {and} \bibinfo{person}{Jitao
  Sang}.} \bibinfo{year}{2020}\natexlab{}.
\newblock \showarticletitle{Towards accuracy-fairness paradox: Adversarial
  example-based data augmentation for visual debiasing}. In
  \bibinfo{booktitle}{\emph{Proceedings of the 28th ACM International
  Conference on Multimedia}}. \bibinfo{pages}{4346--4354}.
\newblock


\bibitem[\protect\citeauthoryear{Zhao, Wang, Yatskar, Cotterell, Ordonez, and
  Chang}{Zhao et~al\mbox{.}}{2019}]%
        {zhao2019gender}
\bibfield{author}{\bibinfo{person}{Jieyu Zhao}, \bibinfo{person}{Tianlu Wang},
  \bibinfo{person}{Mark Yatskar}, \bibinfo{person}{Ryan Cotterell},
  \bibinfo{person}{Vicente Ordonez}, {and} \bibinfo{person}{Kai-Wei Chang}.}
  \bibinfo{year}{2019}\natexlab{}.
\newblock \showarticletitle{Gender Bias in Contextualized Word Embeddings}. In
  \bibinfo{booktitle}{\emph{Proceedings of the 2019 Conference of the North
  American Chapter of the Association for Computational Linguistics: Human
  Language Technologies}}, Vol.~\bibinfo{volume}{1}.
\newblock


\bibitem[\protect\citeauthoryear{Zhao, Wang, Yatskar, Ordonez, and Chang}{Zhao
  et~al\mbox{.}}{2018}]%
        {zhao2018gender}
\bibfield{author}{\bibinfo{person}{Jieyu Zhao}, \bibinfo{person}{Tianlu Wang},
  \bibinfo{person}{Mark Yatskar}, \bibinfo{person}{Vicente Ordonez}, {and}
  \bibinfo{person}{Kai-Wei Chang}.} \bibinfo{year}{2018}\natexlab{}.
\newblock \showarticletitle{Gender Bias in Coreference Resolution: Evaluation
  and Debiasing Methods}. In \bibinfo{booktitle}{\emph{Proceedings of the 2018
  Conference of the North American Chapter of the Association for Computational
  Linguistics: Human Language Technologies, Volume 2 (Short Papers)}}.
  \bibinfo{pages}{15--20}.
\newblock


\bibitem[\protect\citeauthoryear{Zhu, Park, Isola, and Efros}{Zhu
  et~al\mbox{.}}{2017}]%
        {zhu2017unpaired}
\bibfield{author}{\bibinfo{person}{Jun-Yan Zhu}, \bibinfo{person}{Taesung
  Park}, \bibinfo{person}{Phillip Isola}, {and} \bibinfo{person}{Alexei~A
  Efros}.} \bibinfo{year}{2017}\natexlab{}.
\newblock \showarticletitle{Unpaired image-to-image translation using
  cycle-consistent adversarial networks}. In
  \bibinfo{booktitle}{\emph{Proceedings of the IEEE international conference on
  computer vision}}. \bibinfo{pages}{2223--2232}.
\newblock


\end{thebibliography}

\end{document}